%% file: posetrack_paper.tex
\documentclass[10pt,twocolumn,letterpaper]{article}

\usepackage{cvpr}
\usepackage{times}
\usepackage{epsfig}
\usepackage{graphicx}
\usepackage{amsmath}
\usepackage{amssymb}
\usepackage{tabularx}
\usepackage{booktabs}
\usepackage{subcaption}
\usepackage{sistyle}
\SIthousandsep{,}

\newcommand{\myparagraph}[1]{\vspace{0.1em}\noindent\textbf{#1}}

\usepackage[breaklinks=true,bookmarks=false, hidelinks]{hyperref}

\cvprfinalcopy % *** Uncomment this line for the final submission

\def\cvprPaperID{1022} % *** Enter the CVPR Paper ID here

\newcommand{\ProTracker}{ProTracker~\cite{ProTracker} \xspace}

\newcommand{\BUTD}{BUTD~\cite{BUDT}\xspace}

\newcommand{\SOPTPT}{SOPT-PT~\cite{SOPT-PT}\xspace}

\newcommand{\MLLab}{ML-LAB~\cite{ML-LAB}\xspace}

\newcommand{\ICG}{ICG~\cite{ICG}\xspace}

\newcommand{\BUTDS}{BUTDS~\cite{BUDT}\xspace}

\newcommand{\SSDHG}{SSDHG\xspace}

\newcommand{\MPI}{$^1$}
\newcommand{\UBO}{$^2$}

\newcommand{\AMZ}{$^3$}
\newcommand{\GOG}{$^4$}

\title{PoseTrack: A Benchmark for Human Pose Estimation and Tracking
\vspace{-3mm}}
\author{
Mykhaylo Andriluka\GOG$^{,**}$\quad
Umar Iqbal\UBO \quad
Eldar Insafutdinov\MPI \quad
Leonid Pishchulin\MPI\\
Anton Milan\AMZ$^{,*}$ \quad
Juergen Gall\UBO \quad
Bernt Schiele\MPI \\[.5em]
\MPI MPI for Informatics, Saarbr\"ucken, Germany\\
\UBO Computer Vision Group, University of Bonn, Germany\\
\AMZ Amazon Research\\
\GOG Google Research\\
}

\begin{document}
\twocolumn[{%
\renewcommand\twocolumn[1][]{#1}%
\maketitle
\begin{center}
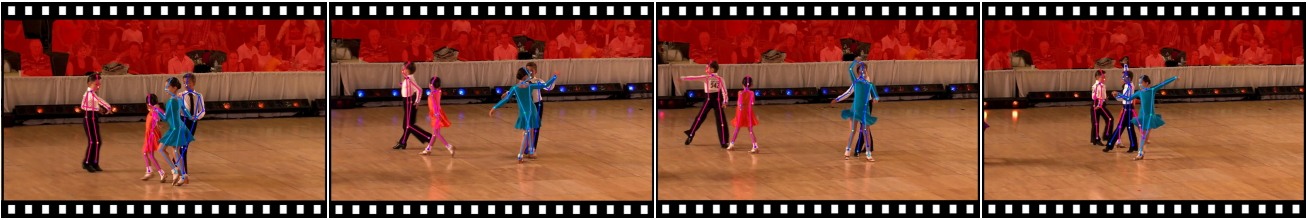

	\vspace{-3mm}
    \centering
    \input{figure_teaser}
    %\includegraphics[width=.8\textwidth,height=5cm]{example-image}
    \vspace{-0.25cm} \captionof{figure}{Sample video from our
      benchmark. We select sequences that represent crowded scenes
      with multiple articulated people engaging in various dynamic
      activities and provide dense annotations of person tracks, body
      joints and ignore regions.}
\end{center}%
}]

%\maketitle
%\thispagestyle{empty}

%%%%%%%%% ABSTRACT
\begin{abstract}
\input{abstract}
\end{abstract}

\renewcommand*{\thefootnote}{\fnsymbol{footnote}}
\footnotetext{$^*$This work was done prior to joining Amazon.}
\renewcommand*{\thefootnote}{\arabic{footnote}}

\renewcommand*{\thefootnote}{\fnsymbol{footnote}}
\footnotetext{$^{**}$This work was done prior to joining Google.}
\renewcommand*{\thefootnote}{\arabic{footnote}}

\input{intro}

\input{related_work}

\input{dataset}
\input{methods}

\input{experiments}

\input{conclusion}
\input{acknowledgements}
{\small
\bibliographystyle{ieee}
\bibliography{egbib,refs-short,anton-ref,biblio}
}

\end{document}

%% file: figure_teaser.tex
\tabcolsep 0.5pt
%\begin{figure*}[t]
\vspace{-0.5cm}
\centering
  \begin{tabular}{c c c c}

	\includegraphics[height=0.1635\linewidth]{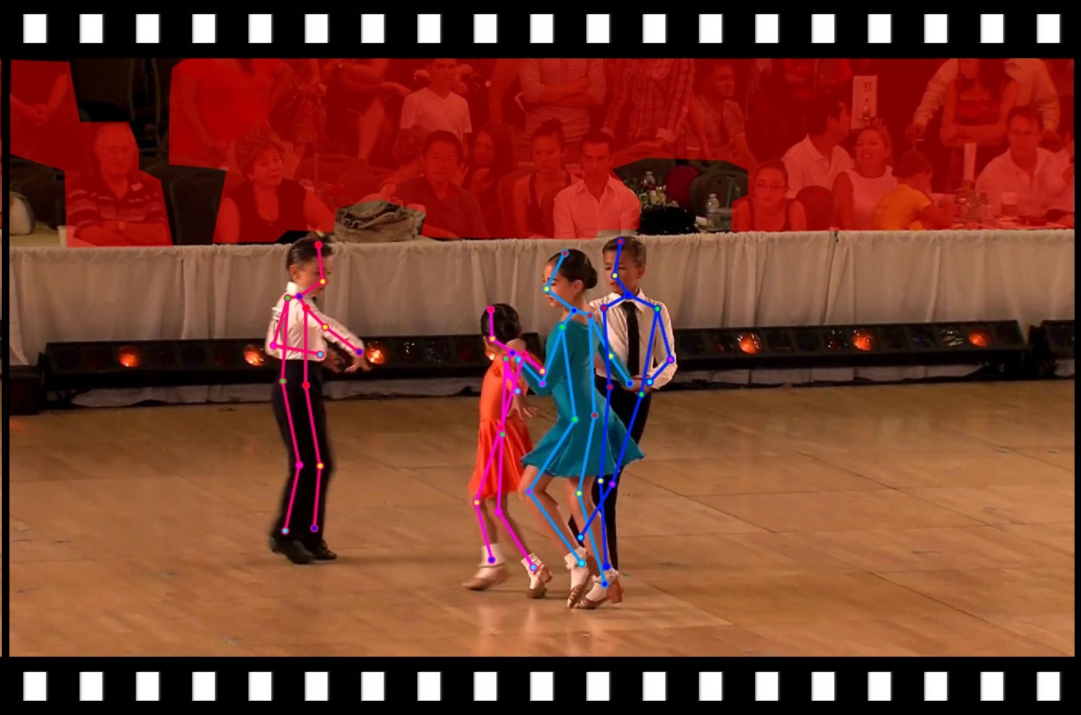}    & 
	
	\includegraphics[height=0.1635\linewidth]{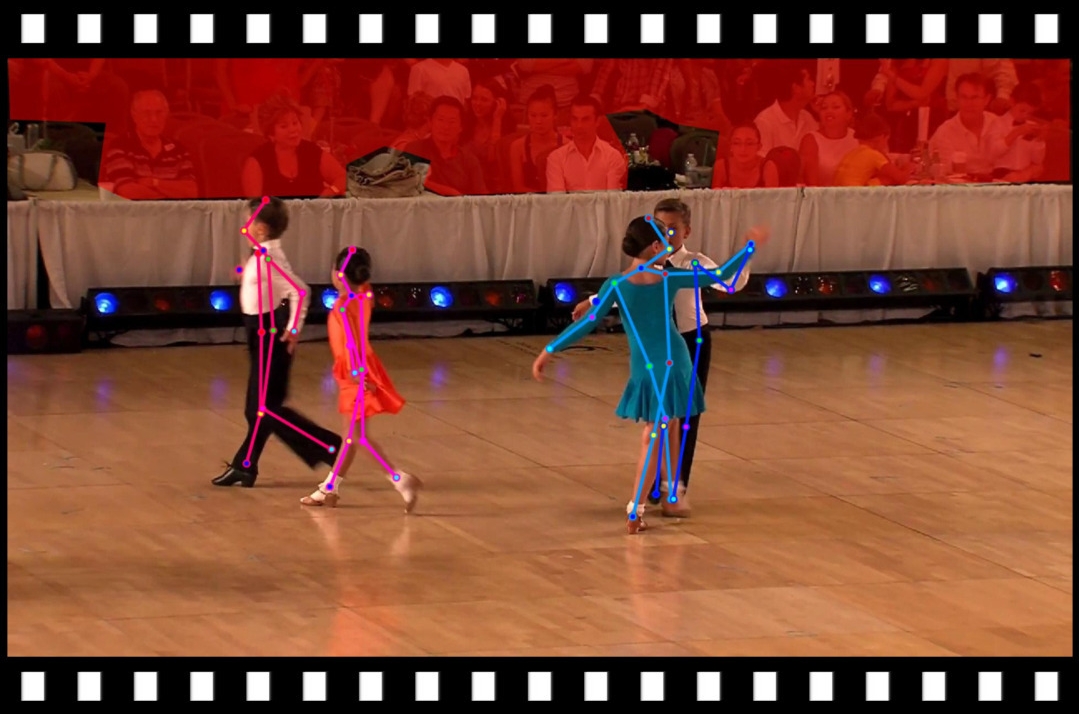}    &  
	
	\includegraphics[height=0.1635\linewidth]{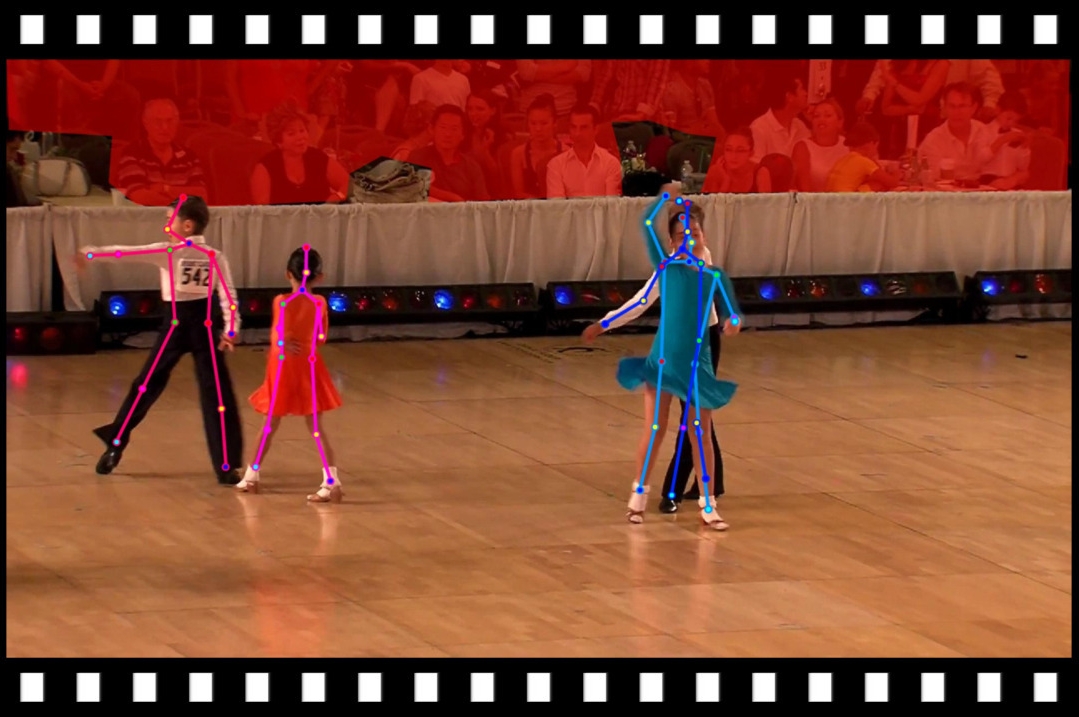} & 
	
	\includegraphics[height=0.1635\linewidth]{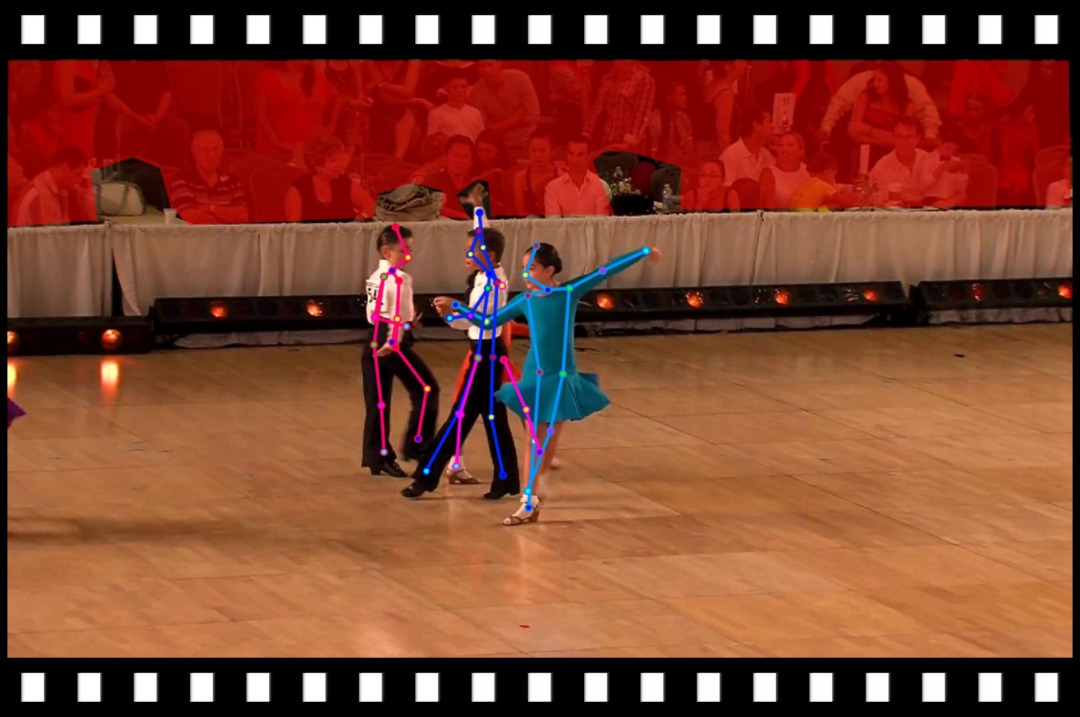} \\

  \end{tabular} 

	%\caption{\todo{TEASER description}}
	%\label{fig:teaser}
%\end{figure*}

%% file: abstract.tex
% Human poses and motions are important cues for representing videos of
% people at a higher level of abstraction. 

Existing systems for video-based pose estimation and tracking struggle
 to perform well on realistic videos with multiple people and often fail to
 output body-pose trajectories consistent over time. To address
this shortcoming this paper introduces 
% Human poses and motions are important cues for analysis of videos with people.
% There is strong evidence that representations based on body pose are highly
% effective for a variety of tasks such as activity recognition, content retrieval
% and social signal processing. 
% In this paper we propose 
PoseTrack which is
% , we aim to further advance the state of the art by establishing 
a new large-scale benchmark for  % , PoseTrack, for
video-based human pose estimation and articulated tracking.
% , and bringing together the community of researchers working on visual human analysis.
 Our new benchmark encompasses three tasks focusing on
i) single-frame multi-person pose estimation, ii) multi-person pose
estimation in videos, and iii) multi-person articulated tracking. To establish the benchmark, we
collect, annotate and release a new %large-scale benchmark
dataset that features videos with multiple people labeled with person
tracks and articulated pose. A public
centralized evaluation server is provided to allow the research community to
evaluate on a held-out test set. 
Furthermore, we conduct an extensive experimental study on recent approaches
to articulated pose tracking and provide analysis of the 
 strengths and weaknesses of the state of the art.
We envision that the proposed
benchmark will stimulate productive research both by providing a large
and representative training dataset as well as providing a platform to
objectively evaluate and compare the proposed methods. The benchmark is freely accessible at \url{https://posetrack.net/}.

\vspace{-5mm}
% Human poses and motions are seen as important measurements for representing videos of people at a
% higher level of abstraction, and there is a strong evidence that pose-based representations are
% effective for a variety of tasks such as activity recognition and content retrieval. In this
% workshop we aim to achieve the following goals: (1) establish a new high-quality benchmark for
% video-based human pose estimation and articulated tracking, and (2) bring together the community of
% researchers working on visual human analysis and articulated pose estimation. To that end we collect
% and make publicly available a new large-scale benchmark dataset that features videos with people
% labeled with person tracks, articulated pose and occlusion labels. The workshop will be organized
% around a challenge with three competition tracks focusing on single-frame pose estimation, pose
% estimation and video, and multi-person articulated tracking. Evaluation server will be setup to
% allow participants to evaluate on a held-out test set. We envision that the proposed benchmark will
% stimulate productive research both by providing a large and representative training dataset as well
% as providing a platform to objectively evaluate and compare the proposed methods.

%% file: intro.tex
\section{Introduction}

Human pose estimation has recently made significant progress on the tasks
of single person pose estimation in individual
frames~\cite{Toshev:2014:DHP,tompson14nips,Tompson:2015:EOL,carreira16cvpr,wei16cvpr, hu2016bottom,insafutdinov16eccv, newell16eccv, bulat2016human, rafi2016bmvc}
and videos~\cite{Pfister15,Charles16,iqbal17fg,gkioxari16eccv} as well as multi-person
pose estimation in monocular
images~\cite{pishchulin16cvpr,insafutdinov16eccv,Iqbal_ECCVw2016,cao16arxiv,papandreou17arxiv}. This
progress has been facilitated by the use of deep learning-based
architectures \cite{Simonyan14c,he2015deep} and by the availability of
large-scale benchmark datasets such as ``MPII Human
Pose''~\cite{andriluka14cvpr} and ``MS
COCO''~\cite{lin14eccv}. Importantly, these benchmark datasets not
only have provided extensive training sets required for training of
deep learning based approaches, but also established detailed metrics
for direct and fair performance comparison across numerous competing
approaches.

Despite significant progress of single frame based multi-person pose
estimation, the problem of \emph{articulated multi-person body joint
tracking} in monocular video remains largely unaddressed.  Although
there exist training sets for special scenarios, such as
sports~\cite{zhang2013actemes,Jhuang:ICCV:2013} and upright frontal
people~\cite{Charles16}, these benchmarks focus on \textit{single
  isolated individuals} and are still limited in their scope and
variability of represented activities and body motions. In this
work, we aim to fill this gap by establishing a new large-scale,
high-quality benchmark for video-based multi-person pose estimation
and articulated tracking. 
% To that end, we collect a videos, annotate them in an accurate and unified manner, and release a new
% large-scale dataset and evaluation platform for the task at hand.

Our benchmark is organized around three related tasks focusing on
single-frame multi-person pose estimation, multi-person pose estimation in video, and multi-person
articulated tracking. While the main focus of the dataset is on multi-person articulated tracking,
progress in the single-frame setting will inevitably improve overall tracking quality. We thus make
the single frame multi-person setting part of our evaluation procedure. In order to enable timely and scalable
evaluation on the held-out test set, we provide a centralized evaluation server. We strongly believe
that the proposed benchmark will prove highly useful to drive the research
forward by focusing on remaining limitations of the state of the art. 

To sample the initial interest of the computer vision community and to obtain
early feedback we have organized a  workshop and a competition at
ICCV'17\footnote{\url{https://posetrack.net/workshops/iccv2017/}}.
%one of the recent computer vision meetings. 
We obtained largely positive feedback from
the twelve teams that participated in the competition. We incorporate some of
this feedback into this paper. In addition we analyze the currently best
performing approaches and highlight the common difficulties for pose estimation
and articulated tracking.

%% file: related_work.tex
\section{Related Datasets}

\input{figure_examples}

\input{table_datasets}

The commonly used publicly available datasets for evaluation of 2D
human pose estimation are summarized in Tab.~\ref{tab:datasets}. 
% We
% state the total number of annotated body poses, availability of video pose
% labels and multiple annotated persons per frame, and types of data.
The table is split into blocks of single-person single-frame, single-person video,
multi-person single-frame, and multi-person video data.

The most popular benchmarks to date for evaluation of single person
pose estimation are 
``LSP''~\cite{johnson10bmvc} (+ ``LSP Extended''~\cite{johnson11cvpr}), 
``MPII Human Pose (Single Person)''~\cite{andriluka14cvpr} and MS COCO Keypoints
Challenge~\cite{lin14eccv}. LSP and LSP Extended datasets focus
on sports scenes featuring a few sport types. Although a combination of
both datasets results in \num{11000} training poses, the evaluation set of
\num{1000} is rather small. FLIC~\cite{sapp13cvpr} targets a simpler task
of upper body pose estimation of frontal upright individuals in
feature movies. In contrast to LSP and FLIC datasets, MPII
Single-Person benchmark covers a much wider variety of everyday human
activities including various recreational, occupational and
household activities and consists of over \num{26000} annotated poses
with \num{7000} poses held out for evaluation. Both benchmarks focus on
single person pose estimation and provide rough location scale of
a person in question. In contrast, our dataset addresses a much more
challenging task of body tracking of multiple highly articulated
individuals where neither the number of people, nor their locations or
scales are known.

The single-frame multi-person pose estimation setting was introduced
in~\cite{eichner10eccv} along with ``We Are Family (WAF)'' dataset.
While this benchmark is an important step towards more challenging
multi-person scenarios, it focuses on a simplified setting of upper
body pose estimation of multiple upright individuals in group photo
collections. The ``MPII Human Pose (Multi-Person)''
dataset~\cite{andriluka14cvpr} has significantly advanced the multi-person
pose estimation task in terms of diversity and difficulty of
multi-person scenes that show highly-articulated people involved in
hundreds of every day activities. More recently, MS COCO Keypoints
Challenge~\cite{lin14eccv} has been introduced %as an attempt 
to provide a new large-scale benchmark for single frame based
multi-person pose estimation. 
% While it contains over \num{100000}
% annotated poses it includes many images of non-challenging 
% isolated individuals. 
All these datasets are only limited to single-frame based body pose estimation.
In contrast, our dataset also focuses on a more challenging task of multi-person
pose estimation in video sequences containing highly articulated people in dense
crowds. This not only requires annotations of body keypoints, but also a unique
identity for every person appearing in the video.  Our dataset is based on the MPII
Multi-Person benchmark, from which we select a subset of key frames and for each
key frame include about five seconds of video footage centered on the key
frame. We provide dense annotations of video sequences with person tracking and
body pose annotations. Furthermore, we adapt a completely unconstrained
evaluation setup where the scale and location of the persons is completely
unknown. This is in contrast to MPII dataset that is restricted to evaluation on
group crops and provides rough group location and scale. Additionally, we
provide ignore regions to identify the regions containing very large crowds of
people that are unreasonably complex to annotate.

Recently, \cite{Iqbal:2017:CVPR} and \cite{insafutdinov17cvpr} also 
provided datasets for multi-person pose estimation in videos. However, both are 
at a very small scale. \cite{Iqbal:2017:CVPR} provides only 60 videos with most 
sequences containing only $41$ frames, and \cite{insafutdinov17cvpr} 
provides $30$ videos containing only $20$ frames each. While these datasets 
make a first step toward solving the problem at hand, they are certainly 
not enough to cover a large range of real-world scenarios and to learn stronger 
pose estimation models. We on the other hand establish a large-scale benchmark
with a much broader variety and an open evaluation setup. 
The proposed dataset contains over \num{150000} annotated poses and over \num{22000} 
labeled frames.

Our dataset is complementary to recent video datasets, such as J-HMDB
\cite{Jhuang:ICCV:2013}, Penn Action~\cite{zhang2013actemes} and YouTube
Pose~\cite{Charles16}. Similar to these datasets, we provide dense
annotations of video sequences. However, in contrast to
\cite{Jhuang:ICCV:2013,zhang2013actemes,Charles16} that focus on single
isolated individuals we target a much more challenging task of multiple
people in dynamic crowded scenarios. In contrast to YouTube Pose that
focus on frontal upright people, our dataset includes a wide variety of
body poses and motions, and captures people at different scales from a
wide range of viewpoints. In contrast to sports-focused Penn Action
and J-HMDB that focuses on a few simple actions, the proposed dataset
captures a wide variety of everyday human activities while being at least 3x
larger compared to J-HMDB.

%and provides more images and a wider
%coverage of activities (491 in our dataset vs. 21 in J-HMDB), whereas
%J-HMDB provides densely annotated image sequences and larger number of
%videos for each activity. 

Our dataset also addresses a different set of challenges compared to
the datasets such as ``HumanEva'' \cite{Sigal:2010:HSV} and
``Human3.6M'' \cite{Ionescu:2014:H36} that include images and 3D poses
of people but are captured in controlled indoor environments,
whereas our dataset includes real-world video sequences but provides
2D poses only.

%% file: figure_examples.tex
\tabcolsep 0.5pt
\begin{figure}[t]
  \centering
  \begin{tabular}{c c}
    \includegraphics[width=0.5\linewidth]{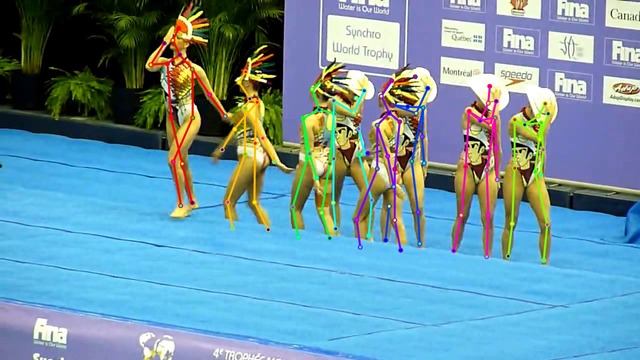}  &
    \includegraphics[width=0.5\linewidth]{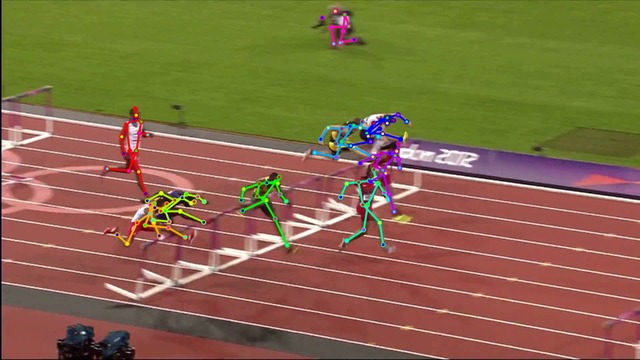} \\
    \includegraphics[width=0.5\linewidth]{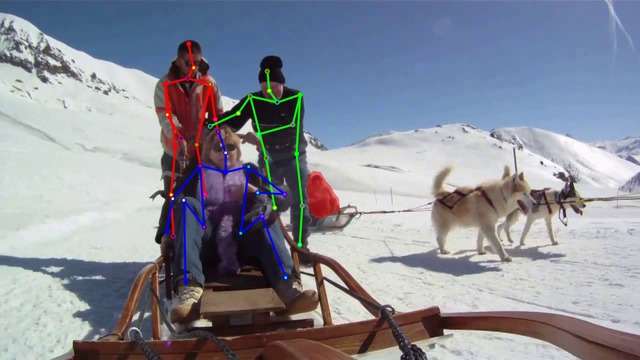} &
    \includegraphics[width=0.5\linewidth]{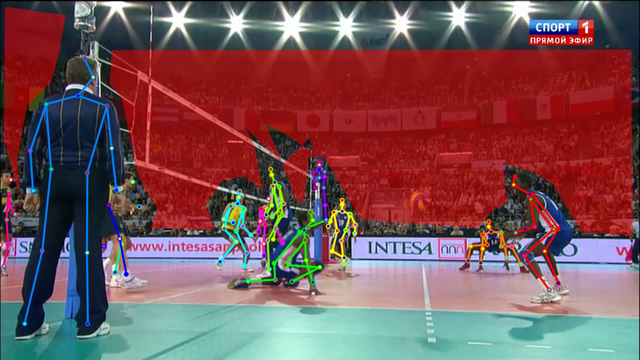} \\
  \end{tabular} 
  \caption{Example frames and annotations from our dataset.}
    \label{fig:annotation_examples}
    \vspace{-0.3cm} 
\end{figure}

%% file: table_datasets.tex
\renewcommand{\arraystretch}{1.2}% Tighter
\tabcolsep 1.5pt

\begin{table}
\centering
\resizebox{\linewidth}{!} {%
% \scriptsize
\begin{tabular}{lrccr}
\toprule
Dataset                                       & \# Poses  & Multi-     & Video-labeled & Data type          \\
                                              &           & person     & poses         &                    \\
\midrule
LSP~\cite{johnson10bmvc}                      &   2,000   &            &               & sports (8 act.)    \\
LSP Extended~\cite{johnson11cvpr}             &  10,000   &            &               & sports (11 act.)   \\
MPII Single Person~\cite{andriluka14cvpr}     &  26,429   &            &               & diverse (491 act.) \\
FLIC~\cite{sapp13cvpr}                        &   5,003   &            &               & feature movies     \\
FashionPose~\cite{dantone13cvpr}              &   7,305   &            &               & fashion blogs      \\
\midrule
We are family~\cite{eichner10eccv}            &  3,131     & \checkmark &               & group photos       \\
MPII Multi-Person~\cite{andriluka14cvpr}      &  14,993   & \checkmark &               & diverse (491 act.) \\
MS COCO Keypoints~\cite{lin14eccv}            & 105,698   & \checkmark &               & diverse            \\
\midrule
Penn Action~\cite{zhang2013actemes}                  & 159,633   &            & \checkmark    & sports (15 act.)   \\
JHMDB~\cite{Jhuang:ICCV:2013}                 &  31,838   &            & \checkmark    & diverse (21 act.)  \\
YouTube Pose~\cite{Charles16}                 &   5,000   &            & \checkmark    & diverse            \\
Video Pose 2.0~\cite{sapp11cvpr}              &   1,286   &            & \checkmark    & TV series          \\
\midrule
Multi-Person PoseTrack~\cite{Iqbal:2017:CVPR} &  16,219  & \checkmark & \checkmark    & diverse            \\
{\bf Proposed}                                     & {\bf 153,615}   & {\bf \checkmark} & {\bf \checkmark}    & {\bf diverse}            \\
\bottomrule
\end{tabular}
% \vspace{0.2cm}
}
\caption{Overview of publicly available datasets for articulated
  human pose estimation in single frames and video. For each dataset
  we report the number of annotated poses, availability of video pose
  labels and multiple annotated persons per frame, as well as types of data.}
\label{tab:datasets}
    \vspace{-0.3cm} 
\end{table}

%% file: dataset.tex
\section{The PoseTrack Dataset and Challenge}
\label{sec:dataset}

%The new large-scale video dataset for articulated human pose tracking in the wild will be the central component introduced and discussed in the workshop. Building on the organizers' experience with creating and maintaining datasets for pose estimation (MPIIHumanPose) and tracking (MOTChallenge), PoseTrack will be established as the new standard benchmark for human pose estimation and tracking. To ensure a fair competition and to discourage overfitting, only the annotations for theThe new large-scale video dataset for articulated human pose tracking in the wild will be the central component introduced and discussed in the workshop. Building on the organizers' experience with creating and maintaining datasets for pose estimation (MPIIHumanPose) and tracking (MOTChallenge), PoseTraack will be established as the new standard benchmark for human pose estimation and tracking. To ensure a fair competition and to discourage overfitting, only the annotations for the training set will be publicly released.
% training set will be publicly released.

% In this work we introduce a new large-scale video dataset and challenge for 
% articulated human pose tracking in the wild.
We will now provide the details on data collection and the annotation process, as well as the established evaluation procedure.
 We build on and extend the newly
 introduced datasets for pose tracking in the wild \cite{insafutdinov17cvpr,Iqbal:2017:CVPR}. To that end, we use the raw videos provided by the popular MPII
Human Pose dataset. For each frame in MPII Human Pose dataset we include $41-298$ neighboring 
frames from the corresponding raw videos, and then select sequences that represent 
crowded scenes with multiple articulated people engaging in various dynamic activities. 
The video sequences are chosen such that they contain a large amount of body motion and body pose and appearance variations. They also contain severe body part occlusion and truncation, \ie, due to occlusions with other people or objects, persons often disappear partially or completely and re-appear again. The scale of the persons also varies across the video due to the movement of persons and/or camera zooming. Therefore, the number of visible persons and body parts also varies across the video. 

%Test 65 - 298
%Train 41 - 151
%Validation 65 - 151

\subsection{Data Annotation} 
We annotated the selected video sequences with person locations,
identities, body pose and ignore regions. The annotations were
performed in four steps. First, we labeled ignore regions to
exclude crowds and people for which pose can not be reliably determined due to
poor visibility. Afterwards, the head bounding boxes
for each person across the videos were annotated and a track ID was 
assigned to each person. The head bounding boxes provide an estimate
of the absolute scale of the person required for evaluation. We assign
a unique track ID to each person appearing in the video until the
person moves out of the camera field-of-view.  Note that each video in our
dataset might contain several shots. We do not maintain track ID between shots
and same person might get different track ID if it reappears in another shot. 
%
% Since a video can contain multiple video shots, we found that person re-identification
% between different shots can sometimes be very difficult even for the
% human annotators. We therefore, assign a new ID to a person if they
% reappear in the video or appear in multiple shots. Since, in this work
% we do not target person re-identification, having different ID for the
% same person in different shots is not very crucial. 
%
Poses for each person track are then annotated in the entire video. We annotate
15 body parts for each body pose including \textit{head, nose, neck, shoulders,
  elbows, wrists, hips, knees and ankles}. All pose annotations were performed
using the VATIC tool~\cite{vodrick12ijcv} that allows to speed-up annotation by
interpolating between frames. We chose to skip annotation of the body joints
that can not be reliably localized by the annotator due to strong occlusion or
difficult imaging conditions. This has proven the be a faster alternative to
requiring annotators to guess the location of the joint and/or marking it as
occluded. Fig.~\ref{fig:annotation_examples} shows example frames from the dataset. Note
the variability in appearance and scale, and complexity due to substantial number of
people in close proximity.
\input{fig_stats}

% All interpolation errors were manually rectified by visualizing the annotations multiple times. Finally, we performed
% two additional iterations of data cleaning to remove any annotation errors and
% obtain high quality annotations. All annotations were performed by in-house
% workers. The annotators were provided with a clearly defined protocol, detailed
% instructions containing several examples and video tutorials, and the authors
% were always accessible to resolve any annotation
% ambiguity.
  
Overall, the dataset contains $550$ video sequences with \num{66374}
frames. We split them into $292$, $50$, $208$ videos for training, validation
and testing, respectively. The split follows the original split of the MPII
Human Pose dataset making it possible to train a model on the MPII Human Pose
and evaluate on our test and validation sets. 
% Further statistics of each subset
% are shown in Tab.~\ref{tab:dataset_stats}.

The length of the majority of the sequences in our dataset ranges between
$41$ and $151$ frames. The sequences correspond to about five seconds of
video. Differences in the sequence length are due to variation in the frame
rate of the videos. A few sequences in our dataset are longer than five seconds
with the longest sequence having $298$ frames. For each sequence in our
benchmark we annotate the $30$ frames in the middle of the sequence. In
addition, we densely annotate validation and test sequences with a step of four
frames. The rationale behind this annotation strategy is that we aim to evaluate both
smoothness of body joint tracks as well as ability to track body joints over
longer number of frames. We did not densely annotate the training set to save
the annotation resources for the annotation of the test and validation set.
% In order to reduce the annotation effort while also including more diverse
% scenarios, we annotated training and validation/testing videos in different
% manners. 
% The length of the training videos ranges between $41$-$151$ frames and
% we densely annotate $30$ frames from the center of the video. Whereas, the
% number of frames in validation/testing videos ranges between $65$ to $298$
% frames. 
% In this case, we densely annotated $30$ frames around the keyframe from
% MPII Pose dataset and afterwards annotate every fourth frame.  
% This strategy allows us to evaluate more diverse body pose articulations and also long-range
% articulated tracking, while having significantly fewer annotations. 
In total, we provide around \num{23000} labeled frames with \num{153615}
pose annotations. To the best of our knowledge this makes PoseTrack the largest multi-person pose
estimation and tracking dataset released to date.  In Fig.~\ref{fig:datastats} we show additional
statistics of the validation and test sets of our dataset. The plots show the
distributions of the number of people per frame and per video, the track length and
people sizes measured by the head bounding box. Note that substantial portion of
the videos has a large number of people as shown in the plot on the
top-right. The abrupt fall off in the plot of the track length in the
bottom-left is due to fixed length of the sequences included in the dataset.
% The already available data in \cite{Iqbal:2017:CVPR,
%    Insafutdinov:2017:CVPR} comprises approximately 100 video sequences
%  totalling around \num{4500} labeled frames and \num{20000} annotated
%  humans. However, in order to be consistent across the dataset and
%  have the same skeleton structure for all annotations, we re-annotated all
%  videos.  
% The testing set remains unpublished to avoid over-fitting,
%  and no information about the test set is revealed including the
%  information about which frames are labeled.

\subsection{Challenges} The benchmark consists of the following challenges:
% \begin{enumerate}

\myparagraph{Single-frame pose estimation.} This task is similar to the ones covered by existing datasets like MPII Pose and MS COCO Keypoints, but on our new large-scale dataset.

\myparagraph{Pose estimation in videos.} The evaluation of this challenge is performed on single frames, however, the data will also include video frames before and after the annotated ones, allowing methods to exploit video information for a more robust single-frame pose estimation.

\myparagraph{Pose tracking.} This task requires to provide temporally consistent poses for all people visible in the videos. Our evaluation include both individual pose accuracy as well as temporal consistency measured by identity switches.
% \end{enumerate}

\subsection{Evaluation Server} We provide an online evaluation server to quantify the performance of different methods on the held-out test set. This will not only prevent over-fitting to the test data but also ensures that all methods are evaluated in the exact same way, using the same ground truth and evaluation scripts, making the quantitative comparison meaningful. Additionally, it can also serve as a central directory of all available results and methods.

\subsection{Experimental Setup and Evaluation Metrics}
\label{sec:metrics}
Since we need to evaluate both the accuracy of multi-person pose
estimation in individual frames and articulated tracking in videos, we
follow the best practices followed in both multi-person pose
estimation \cite{pishchulin16cvpr} and multi-target tracking
\cite{Milan:2016:MOT16}. In order to evaluate whether a body part is
predicted correctly, we use the PCKh (head-normalized probability of
correct keypoint) metric \cite{andriluka14cvpr}, which considers a
body joint to be correctly localized if the predicted location of the
joint is within a certain threshold from the true location. Due to 
large scale variation of people across videos and even within a frame,
this threshold needs to be selected adaptively based on the person's
size. To that end, we follow~\cite{andriluka14cvpr} and use $50\%$ of
the head length where the head length corresponds to $60\%$ of the
diagonal length of the ground-truth head bounding box. Given the joint
localization threshold for each person, we compute two sets of
evaluation metrics, one which is commonly used for evaluating
multi-person pose estimation~\cite{pishchulin16cvpr}, and one from the
multi-target tracking literature~\cite{Yang:2012:CVPR, Choi:2015:ICCV,
  Milan:2016:MOT16} to evaluate multi-person pose tracking. During evaluation we 
ignore all person detections that overlap with the ignore regions. 

\myparagraph{Multi-person pose estimation.}  For measuring frame-wise
multi-person pose accuracy, we use \emph{mean Average Precision} (mAP)
as is done in \cite{pishchulin16cvpr}.
The protocol to evaluate multi-person pose estimation in
\cite{pishchulin16cvpr} requires that the location of a group of
persons and their rough scale is known during evaluation
\cite{pishchulin16cvpr}. This information, however, is almost never
available in realistic scenarios, particularly for videos. We
therefore, propose not to use any ground-truth information during
testing and evaluate the predictions without rescaling or selecting a
specific group of people for evaluation.

\myparagraph{Articulated multi-person pose tracking.}  To evaluate multi-person
pose tracking, we use Multiple Object Tracking (MOT) metrics
\cite{Milan:2016:MOT16} and apply them independently to each of the body
joints. Metrics measuring the overall tracking performance are then obtained by
averarging the per-joint metrics. 
The metrics require predicted body poses with track IDs. First, for each frame,
for each body joint class, distances between predicted and ground-truth locations are computed. 
%Then, predicted track IDs and GT track IDs are taken into account and 
Subsequently predicted and ground-truth locations are matched to each other
by a global matching procedure that minimizes the total assignment distance.
% In the following all pairs of prediction and GT with distances below the PCKh
% threshold are considered during global matching of predicted tracks to GT tracks
% for each of the body joints. Global matching minimizes the total assignment
% distance. 
Finally, Multiple Object Tracker Accuracy (MOTA), Multiple Object
Tracker Precision (MOTP), Precision, and Recall metrics are computed. Evaluation
server reports MOTA metric for each body joint class and average over all body
joints, while for MOTP, Precision, and Recall we report averages only. In the
following evaluation MOTA is used as our main tracking metric.
The source code for the evaluation metrics is publicly available on the
benchmark website.

%% file: fig_stats.tex
\begin{figure}[t]
	\centering
	\includegraphics[height=0.31\columnwidth]{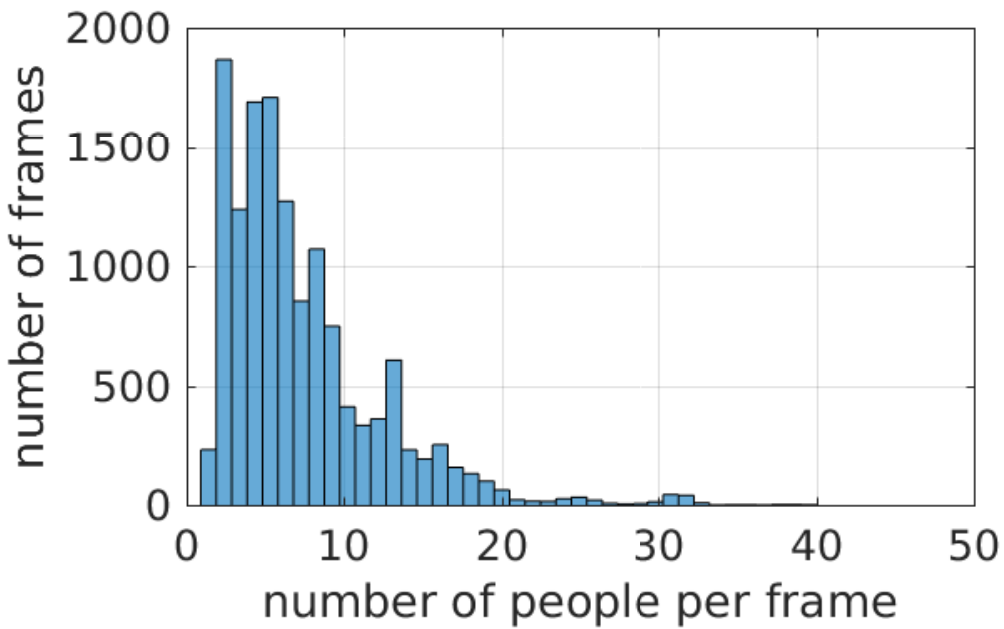} 
	\includegraphics[height=0.31\columnwidth]{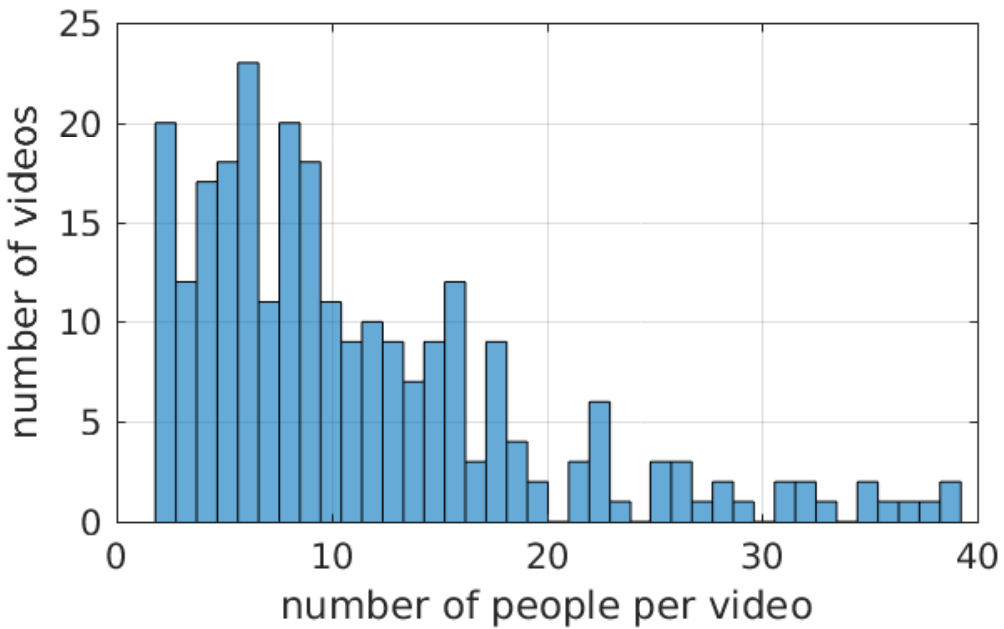} 
	\hfill
	 \\
	\includegraphics[height=0.31\linewidth]{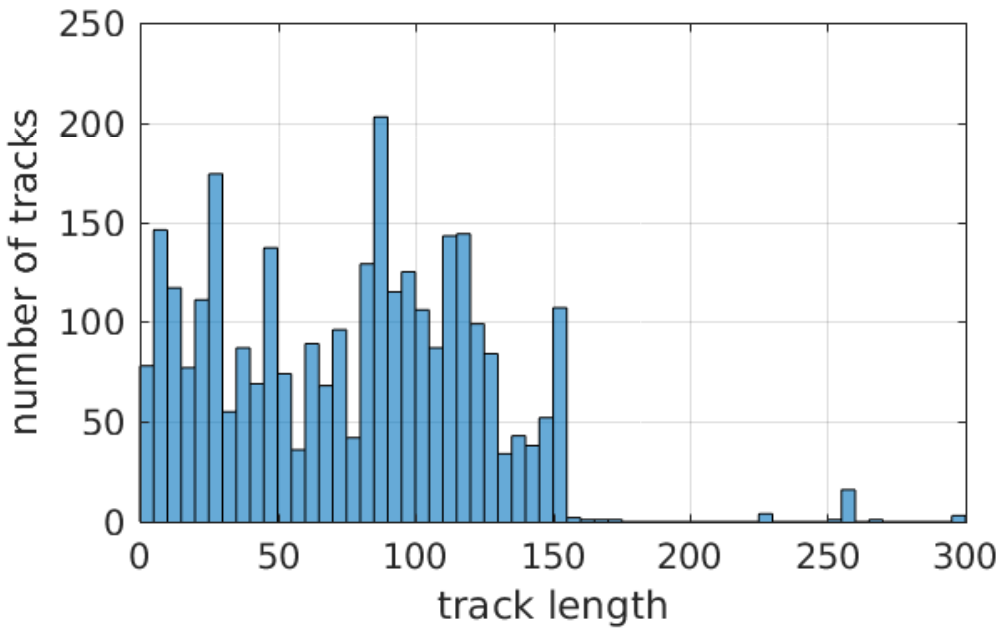} 
	\includegraphics[height=0.31\linewidth]{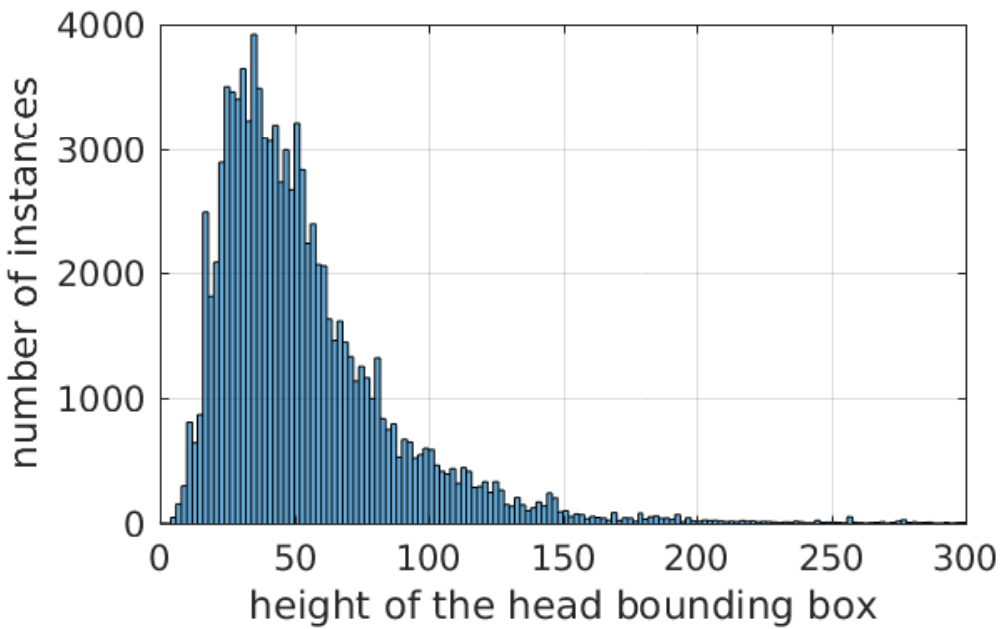} 
	\hfill
	\\
	\includegraphics[height=0.31\linewidth]{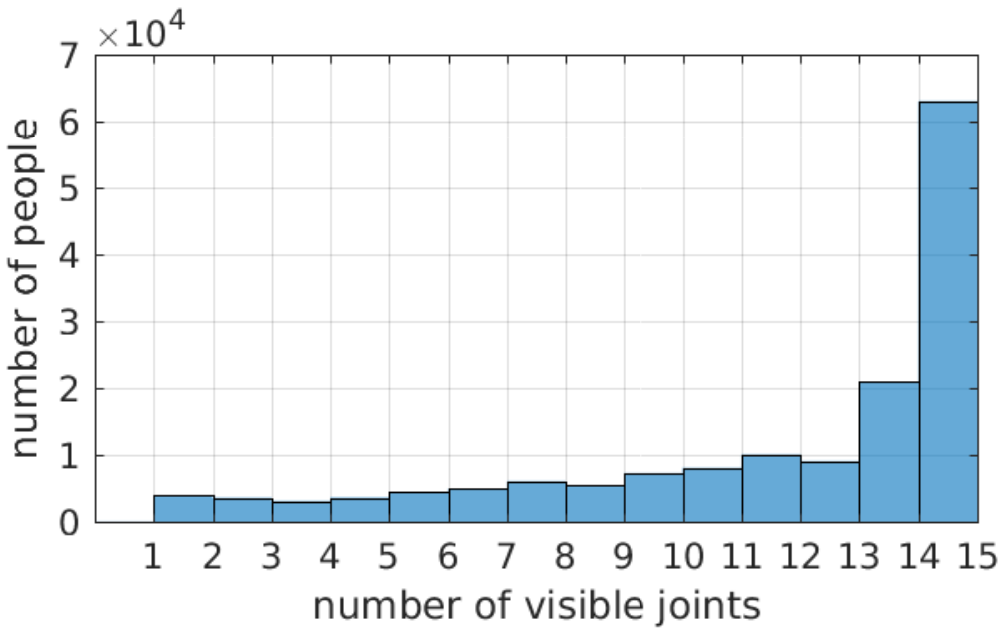} 
	\includegraphics[height=0.31\linewidth]{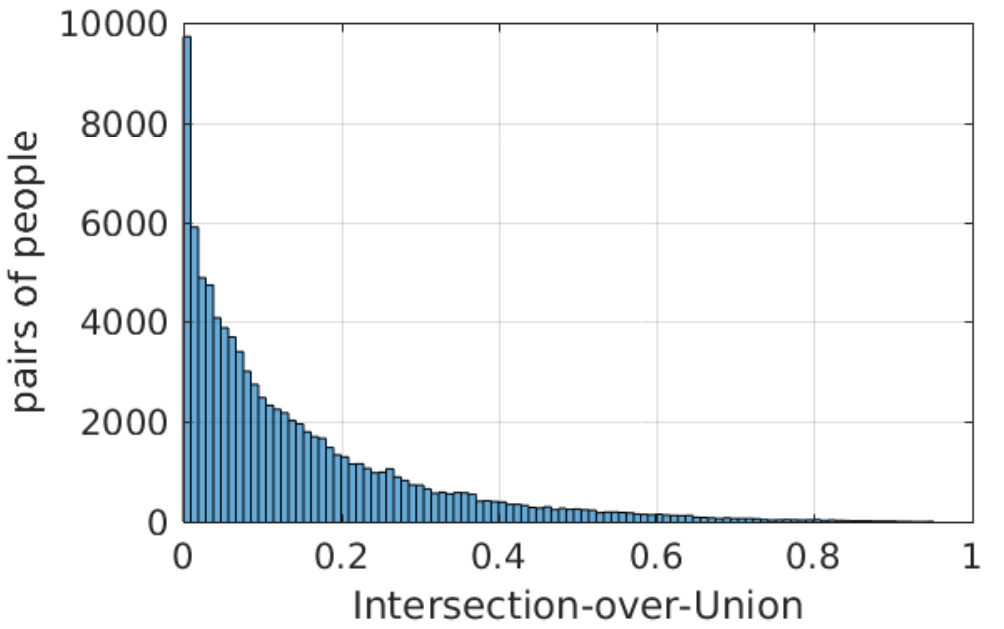} 
	\hfill
	\\
	\caption{Various statistics of the PoseTrack benchmark.}
        \vspace{-0.35cm}
	\label{fig:datastats}
\end{figure}

%% file: methods.tex
\section{Analysis of the State of the Art}
\label{sec:analysis-sota}
\input{table_posetrack_mota.tex}
\input{table_posetrack_map.tex}
\input{table_per_frame_ap_val.tex}
\input{table_tracking_mota_val.tex}

Articulated pose tracking in unconstrained videos is a relatively new topic in
computer vision research. To the best of our knowledge only few approaches for
this task have been proposed in the literature \cite{insafutdinov17cvpr,Iqbal:2017:CVPR}.
Therefore, to analyze the performance of the state of the art on our
new dataset, we proceed in two ways. 

%First, we propose a baseline
%method based on these state-the-art approaches
%\cite{insafutdinov17cvpr,Iqbal:2017:CVPR}. This baseline employs the
%same graph-partitioning formulation as
%\cite{insafutdinov17cvpr,Iqbal:2017:CVPR}, but relies on a person
%detector to localize people prior to pose estimation. This additional
%person detection step is necessary because the original methods do not
%handle well the large variability in people scales present in our new
%dataset. Note that our benchmark includes an order of magnitude more
%sequences compared to the ``MPII Video Pose'' dataset used in
%\cite{insafutdinov17cvpr} and the sequences in our benchmark are about
%five times longer, which makes it computationally expensive to run the
%graph partitioning on the full sequences as in
%\cite{insafutdinov17cvpr} without additional pruning of the hypothesis
%space.
%
First, we propose two baseline methods based on the state-of-the-art approaches \cite{insafutdinov17cvpr,Iqbal:2017:CVPR}. Note that our benchmark includes an order of magnitude more
sequences compared to the datasets used in
\cite{insafutdinov17cvpr,Iqbal:2017:CVPR} and the sequences in our benchmark are about
five times longer, which makes it computationally expensive to run the
graph partitioning on the full sequences as in
\cite{insafutdinov17cvpr, Iqbal:2017:CVPR}. We modify these methods to make them applicable on our proposed dataset. The baselines and corresponding modifications are explained in Sec.~\ref{sec:baseline_methods}.

Second, in order to broaden the scope of our evaluation we organized a  \emph{PoseTrack Challenge} in conjunction with ICCV'17 on our dataset by establishing an online evaluation
server and inviting submissions from the research community. 
%To
%encourage submissions we organized a challenge competition at one of
%the recent major computer vision meetings./
% We keep the names of the
% submitted methods hidden to preserve anonymity of the authors and of
% this submission.
%\footnote{The full references to the submitted
% approaches will be included in the paper upon publication.} 
In the following we consider the top five methods submitted to the online
evaluation server both for the pose estimation and pose tracking
tasks. In Tab.~\ref{tab:posetrack_mota} and \ref{tab:posetrack_map} we
list the best performing methods on each task sorted by MOTA and mAP,
respectively.  In the following we first describe our baselines based
on \cite{insafutdinov17cvpr,Iqbal:2017:CVPR} and then summarize the
main observations made in this evaluation.

\subsection{Baseline Methods}
\label{sec:baseline_methods}

We build the first baseline model following the graph partitioning
formulation for articulated tracking proposed in
\cite{insafutdinov17cvpr}, but introduce two
simplifications that follow \cite{papandreou17arxiv}. First, we rely on
a person detector to establish locations of people in the image and
run pose estimation independently for each person detection. This
allows us to deal with large variation in scale present in our dataset by
cropping and rescaling images to canonical scale prior to pose
estimation. In addition, this also allows us to group together the
body-part estimates inferred for a given detection bounding box. As a
second simplification we apply the model on the level of full body
poses and not on the level of individual body parts as in
\cite{insafutdinov17cvpr,Iqbal:2017:CVPR}. We use a publicly available Faster-RCNN~\cite{ren2015faster} detector from the
TensorFlow Object Detection API~\cite{huang2016speed} for people detection. This detector has been trained on the ``MS COCO'' dataset and uses Inception-ResNet-V2~\cite{szegedy2017inception} for image encoding.
We adopt the DeeperCut CNN architecture from \cite{insafutdinov16eccv} as our
pose estimation method. This architecture is based on the ResNet-101 converted
to a fully convolutional network by removing the global pooling layer and
utilizing atrous (or dilated) convolutions~\cite{chen2016deeplab} to increase
the resolution of the output scoremaps. Once all poses are extracted, we perform
non-maximum suppression based on pose similarity
criteria~\cite{papandreou17arxiv} to filter out redundant person detections.  We
follow the cropping procedure of~\cite{papandreou17arxiv} with the crop size
336x336px. Tracking is implemented as in \cite{insafutdinov17cvpr} by forming
the graph that connects body-part hypotheses in adjacent frames and partitioning
this graph into connected components using an approach from
\cite{Levinkov_2017_CVPR}. We use Euclidean distance between body joints to
derive costs for graph edges. Such distance-based features were found to be
effective in \cite{insafutdinov17cvpr} with additional features adding minimal
improvements at the cost of substantially slower inference.

For the second baseline, we use the publicly available source code of
\cite{Iqbal:2017:CVPR} and replace the pose estimation model with
\cite{cao16arxiv}. We empirically found that the pose estimation model of
\cite{cao16arxiv} is better at handling large scale variations compared to
DeeperCut \cite{insafutdinov16eccv} used in the original paper. % when performing bottom-up multi-person pose estimation. 
We do not make any changes in the graph partitioning algorithm, but reduce the window size to $21$ as compared to $31$ used in the original model. We refer the readers to \cite{Iqbal:2017:CVPR} for more details. 
The goal of constructing these strong baselines is to validate the results submitted to our evaluation server and to allow us to perform additional experiments presented in
Sec.~\ref{sec:baselines}. In the rest of this paper, we refer to them as
ArtTrack-baseline and PoseTrack-baseline respectively. 

\subsection{Main Observations}
% \item 2 components: pose detector + simple tracker, as we show this is sufficient for scenes of moderate complexity 
\myparagraph{Two-stage design.}  The first observation is that all
submissions follow a two-stage tracking-by-detection design. In the first stage,
a combination of person detector and single-frame pose estimation method is used
to estimate poses of people in each frame. The exact implementation of
single-frame pose estimation method varies. Each of the top three articulated
tracking methods builds on a different pose estimation approach (Mask-RCNN
\cite{he17iccv}, PAF \cite{cao16arxiv} and DeeperCut
\cite{insafutdinov16eccv}). On the other hand, when evaluating methods according
to pose estimation metric (see Tab.~\ref{tab:posetrack_map}) three of the top
four approaches build on PAF \cite{cao16arxiv}. The performance still varies
considerably among these PAF-based methods (70.3 for submission \MLLab vs. 62.5
for submission \SOPTPT) indicating that large gains can be achieved within the
PAF framework by introducing incremental improvements.

% PAF \cite{} is polular, but other approaches such as \cite{} and
% \cite{} appear to be competitive. 

% The second component is an independent method
% that links pose estimates from the first stage over time. Here two types of
% approaches are used. At this stage approaches \ProTracker, \SOPTPT and \ICG rely
% on some form of bipatite matching between hypothesis in adjacent frames (either
% based on Hungarian algorithm or greedy assignment). In contrast approaches \BUTD
% and ArtTrack rely on graph-partitioning that jointly finds multiple tracks in
% the entire sequence.

In the second stage the single-frame pose estimates are linked over time.  For
most of the methods the assignment is performed on the level of body poses, not
individual parts. This is indicated in the ``Tracking granularity'' column in
Tab.~\ref{tab:posetrack_mota}. Only submission \BUTD and our PoseTrack-baseline track people
on the level of individual body parts. Hence, most methods establish
correspondence/assembly of parts into body poses on the per-frame level. In
practice, this is implemented by supplying a bounding box of a person and running
pose estimation just for this box, then declaring maxima of the heatmaps as
belonging together. This is suboptimal as multiple people overlap significantly,
yet most approaches choose to ignore such cases (possibly for inference
speed/efficiency reasons). The best performing approach \ProTracker relies on
simple matching between frames based on Hungarian algorithm and matching cost
based on intersection-over-union score between person bounding boxes. None of
the methods is end-to-end in the sense that it is able to directly infer
articulated people tracks from video. We observe that the pose tracking
performance of the top five submitted methods saturates at around 50 MOTA, with
the top four approaches showing rather similar MOTA results (51.8 for
submission \ProTracker vs. 50.6 for submission \BUTD vs. 48.4 for
PoseTrack-baseline vs. 48.1 for ArtTrack-baseline). %, indicating room for improvement on this task.
\input{figure_mota_sorted}

% \item best performance requires pre-training on COCO and MPII Pose that include wider variety of people appearance 
\myparagraph{Training data.} Most submissions found it necessary to combine our
training set with datasets of static images such as COCO and MPII-Pose to obtain
a joint training set with larger appearance variability. The most common
procedure was to pre-train on external data and then fine-tune on our
training set. Our training set is composed of \num{2437} people tracks with \num{61178}
annotated body poses and is complementary to COCO and MPII-Pose which include an
order of magnitude more individual people but do not provide motion
information. We quantify the performance improvement due to training on
additional data in Tab.~\ref{tab:per-frame-ap-val} using our ArtTrack baseline.
Extending the training data with the MPII-Pose dataset improves the performance
considerably (55.5 vs. 68.7 mAP). The combination of our dataset and MPII-Pose
still performs better than MPII-Pose alone (66.4 vs. 68.7) showing that datasets
are indeed complementary.

None of the approaches in our evaluation employs any form of learning
on the provided video sequences beyond simple cross-validation of a few hyperparameters.
 This can be in part due to relatively small
size of our training set. One of the lessons learned from our work on this benchmark
is that creating truly large annotated datasets of articulated pose sequences is
a major challenge. We envision that future work will combine manually labeled
data with other techniques such as transfer learning from other datasets such as
 \cite{Carreira_2017_CVPR}, inferring sequences of poses by propagating
annotations from reliable keyframes \cite{Charles16},
and leveraging synthetic training data as in \cite{varol17b}.

%
%\input{table_per_frame_ap_test.tex} 
% Table~\ref{tab:per-frame-ap-test} summarizes pose estimation results
% on the test set. \todo{comparison with the challenge submissions}.
% Table 3 contains the results of ablation study on pose tracking
% \ref{tab:tracking-mota-val}.
%
\input{figure_easy_sequences.tex}
\vspace{0.1cm}
\input{figure_hard_sequences.tex}

\myparagraph{Dataset difficulty.}  We composed our dataset by including videos
around the keyframes from MPII Human Pose dataset that included several people
and non-static scenes. The rationale was to create a dataset that would be
non-trivial for tracking and require methods to correctly resolve effects such
as person-person occlusions. In Fig.~\ref{fig:mota:all_methods} we visualize
performance of the evaluated approaches on each of the test sequences.  We
observe that test sequences vary greatly with respect to difficulty both for
pose estimation as well as for tracking. \Eg, for the best performing submission
\ProTracker the performance varies from nearly 80 MOTA to a score below zero\footnote{Note that MOTA metric can become negative for example when the number
  of false positives significantly exceeds the number of ground-truth
  targets.}. Note that the approaches mostly agree with respect to the
difficulty of the sequences.  More difficult sequences are likely to require
methods that are beyond simple tracking component based on frame-to-frame
assignment used in the currently best performing approaches. To encourage
submissions that explicitly address challenges in the difficult portions of the
dataset we have defined easy/moderate/hard splits of the data and report results
for each of the splits as well as the full set.

\input{figure_art_complexity}

\myparagraph{Evaluation metrics.} The MOTA evaluation metric has a deficiency in
that it does not take the confidence score of the predicted tracks into
account. As a result achieving good MOTA score requires tuning of the pose
detector threshold so that only confident track and pose hypothesis are supplied
for evaluation. This in general degrades pose estimation performance as measured
by mAP (\cf performance of submission \ProTracker in
Tab.~\ref{tab:posetrack_mota} and \ref{tab:posetrack_map}).  We 
quantify this in Fig.~\ref{tab:tracking-mota-val} for our ArtTrack
baseline. Note that filtering the detections with score below $\tau=0.8$ as
compared to $\tau=0.1$ improves MOTA from 38.1 to 53.4. One potential
improvement to the evaluation metric would be to require that pose tracking
methods assign confidence score to each predicted track as is common for pose
estimation and object detection.  This would allow one to compute a final score as
an average of MOTA computed for a range of track scores. Current pose
tracking methods typically do not provide such confidence scores. We believe that extending
the evaluation protocol to include confidence scores is an important future direction. 

%\item competition entries improve over our own implementation state-of-the-art
% \myparagraph{What is the state-of-the-art performance?}  We observe that the
% performance on the pose tracking challenge of the top five submitted methods saturates at around
% 50 MOTA, with the top three approaches showing rather similar MOTA results (51.8
% for submission \ProTracker vs. 50.6 for submission \BUTD vs. 48.1 for ArtTrack).
% The differences in performance on pose estimation challenge are more pronounced (70.3 mAP for
% submission \MLLab vs. 65.1 mAP for ArtTrack vs. 64.5 mAP for submission
% \BUTDS). 

% We hypothesize that substantial innovation in the method design or much
% larger training sets would be needed to break through the current performance
% values.

%% file: table_posetrack_mota.tex
\tabcolsep 6pt
\begin{table*}[tbp]
 % \scriptsize
  \centering
  \resizebox{\linewidth}{!} {%
  \begin{tabular}{l c c c c c c}
    \toprule
    Submission & Pose model & Tracking model &  Tracking granularity &
                                                                       Additional training data & mAP & MOTA  \\
    %&  &  & & & & & & & CNN & infer.\\
    \midrule
\ProTracker & Mask R-CNN \cite{he17iccv}  & Hungarian & pose-level & COCO  & 59.6  & \textbf{51.8} \\ % CMU
\BUTD  & PAF \cite{cao16arxiv}  & graph partitioning & person-level and part-level & COCO  & 59.2 & 50.6 \\ % The Chinese University of Hong Kong
\SOPTPT & PAF \cite{cao16arxiv}  & Hungarian & pose-level & MPII-Pose + COCO  & 62.5 & 44.6\\ % Hikvision Research Institute
\MLLab & modification of PAF \cite{cao16arxiv}  &  frame-to-frame  assign. & pose-level & MPII-Pose + COCO & \textbf{70.3} & 41.8 \\% Samsung Research Beijing
\ICG & novel single-/multi-person CNN & frame-to-frame assign. & pose-level  & - & 51.2 & 32.0 \\% (Gratz)
    \midrule
ArtTrack-baseline & Faster-RCNN
                                      \cite{huang2016speed} +
                                      DeeperCut
                                      \cite{insafutdinov16eccv}  &
                                                                    graph
                                                                    partitioning
                                             & pose-level & MPII-Pose + COCO &  59.4 & 48.1  \\
PoseTrack-baseline & PAF \cite{cao16arxiv} & graph partitioning & part-level & COCO & 59.4 & 48.4 \\
     %\midrule
    \bottomrule
  \end{tabular}
 % \vspace{0.75em}
 }
 \caption[]{Results of the top five pose tracking models submitted to our evaluation
  server and of our baselines based on \cite{insafutdinov17cvpr} and \cite {Iqbal:2017:CVPR}. Note that mAP for some of the methods might be
  intentionally reduced to achieve higher MOTA (see discussion in text).}
\label{tab:posetrack_mota}
%  \vspace{-1.5em}
\end{table*}

%% file: table_posetrack_map.tex
\tabcolsep 6pt
\begin{table}[tbp]
 % \scriptsize
  \centering
  \resizebox{\linewidth}{!} {%
  \begin{tabular}{l c c c}
    \toprule
    Submission & Pose model & Additional training data & mAP  \\
    %&  &  & & & & & & & CNN & infer.\\
    \midrule

% Samsung Research Beijing
\MLLab & modification of PAF \cite{cao16arxiv} & COCO & \textbf{70.3} \\

% The Chinese University of Hong Kong
\BUTDS & PAF \cite{cao16arxiv} & MPII-Pose + COCO & 64.5 \\ 

% CMU 
\ProTracker & Mask R-CNN \cite{he17iccv} & COCO & 64.1 \\

% Hikvision Research Institue
\SOPTPT & PAF \cite{cao16arxiv} & MPII-Pose + COCO & 62.5 \\

% University of Missouri-Columbia
% Submission \FRACTALNET & - &  COCO & 62.4 \\

% South China University of Technology
\SSDHG & SSD \cite{liu2016ssd} + Hourglass \cite{newell16eccv} & MPII-Pose + COCO & 60.0 \\

\midrule
ArtTrack-baseline & DeeperCut & MPII-Pose + COCO & 65.1 \\
PoseTrack-baseline & PAF \cite{cao16arxiv} & COCO & 59.4 \\
     %\midrule
    \bottomrule
  \end{tabular}
 % \vspace{0.75em}
 }
\caption[]{Results of the top five pose estimation models submitted to our evaluation
  server and of our baselines. The methods are ordered according to mAP. Note that the mAP of ArtTrack and submission \ProTracker is different from Tab.~\ref{tab:posetrack_mota} because the evaluation in this table does not threshold detections by the score.}
\label{tab:posetrack_map}
%  \vspace{-1.5em}
\end{table}

%% file: table_per_frame_ap_val.tex
\tabcolsep 2.5pt
\begin{table}[tbp]
 %\scriptsize
  \centering
  \resizebox{\linewidth}{!} {%
  \begin{tabular}{@{} l c c ccc ccc c c@{}}
    \toprule
    Model& Training Set & Head   & Sho  & Elb & Wri & Hip & Knee & Ank & mAP \\
    %&  &  & & & & & & & CNN & infer.\\
    \midrule
     ArtTrack-baseline  & our dataset        & 73.1 & 65.8 & 55.6 & 47.2 & 52.6 & 50.1 & 44.1 & 55.5 \\
     ArtTrack-baseline & MPII               & 76.4 & 74.4 & 68.0 & 59.4 & 66.1 & 64.2 & 56.6 & 66.4 \\
     ArtTrack-baseline & MPII + our dataset & \textbf{78.7} & \textbf{76.2} & \textbf{70.4} & \textbf{62.3} & \textbf{68.1} & \textbf{66.7} & \textbf{58.4} & \textbf{68.7} \\
     %\midrule
    \bottomrule
  \end{tabular}
 %\vspace{0.75em}
 }
\caption[]{Pose estimation performance (mAP) of our ArtTrack baseline for
  different training sets.
}
\label{tab:per-frame-ap-val}
%  \vspace{-1.5em}
\end{table}

%% file: table_tracking_mota_val.tex
\tabcolsep 1.5pt
\begin{table}[bp]
 \scriptsize
  \centering
  \begin{tabular}{@{} c c ccc ccc c | c c@{}}
    \toprule
    Model& Head   & Sho  & Elb & Wri & Hip & Knee & Ank & Total & mAP \\
    %&  &  & & & & & & & CNN & infer.\\
    \midrule
     ArtTrack-baseline, $\tau=0.1$    & 58.0 & 56.4 & 34.0 & 19.2 & 44.1 & 35.9 & 19.0 & 38.1 & \textbf{68.6} \\ 
     ArtTrack-baseline, $\tau=0.5$    & 63.5 & 62.8 & 48.0 & 37.8 & 52.9 & 48.7 & 36.6 & 50.0 & 66.7 \\ 
     ArtTrack-baseline, $\tau=0.8$    & \textbf{66.2} & \textbf{64.2} & \textbf{53.2} & \textbf{43.7} & \textbf{53.0} & \textbf{51.6} & \textbf{41.7} & \textbf{53.4} & 62.1 \\ 
%\midrule
%     PoseTrack, $\tau=0.1$    & 66.6	& 73.5	& 55.5	& 43.5	& 54.0	& 45.1	& 40.0	& 54.9		& 64.9\\ 
%     PoseTrack, $\tau=0.5$    & 63.0	& 69.8	& 56.2	& 41.7	& 47.8	& 42.9	& 38.1	& 52.1		& 53.9 \\ 
%     PoseTrack, $\tau=0.8$    & 37.4	& 31.1	& 22.5	& 15.4	& 9.6	& 15.7	& 10.3	& 21.4		& 17.9 \\ 
    \bottomrule
  \end{tabular}
% \vspace{0.75em}
\caption[]{Pose tracking performance (MOTA) of ArtTrack baseline for different
  part detection cut-off thresholds $\tau$. 
}
\label{tab:tracking-mota-val}
%  \vspace{-1.5em}
\end{table}

%% file: figure_mota_sorted.tex
% \begin{figure}[t]
% 	\includegraphics[trim=0 0 0 0,clip,width=1.0\linewidth]{figures/mota_sorted} \\
% \vspace{-0.5cm}
% 		\caption{Sequences sorted by average MOTA.}
% 	\label{fig:mota:all_methods}
% \vspace{-0.5cm}
% \end{figure}

\begin{figure*}[!tbp]
%  \center
%  \begin{subfigure}[b]{0.30\textwidth}
%    %\includegraphics[trim=0 0 0 0,clip,width=1.0\linewidth]{figures/mota_sorted} \\
%    \includegraphics[width=\textwidth,height=0.17\textheight]{figures/mota_sorted} \\
%  \end{subfigure}
%  \begin{subfigure}[b]{0.30\textwidth}
%    \includegraphics[width=\textwidth,height=0.17\textheight]{figures/articulation_complexity_AP_plot}
%    %\includegraphics[width=\textwidth]{figures/articulation_complexity_MOTA_plot}
%  \end{subfigure}
%  \begin{subfigure}[b]{0.30\textwidth}
%    \includegraphics[width=\textwidth,height=0.17\textheight]{figures/ap_mota}
%  \end{subfigure}
  \includegraphics[scale=1,width=\textwidth]{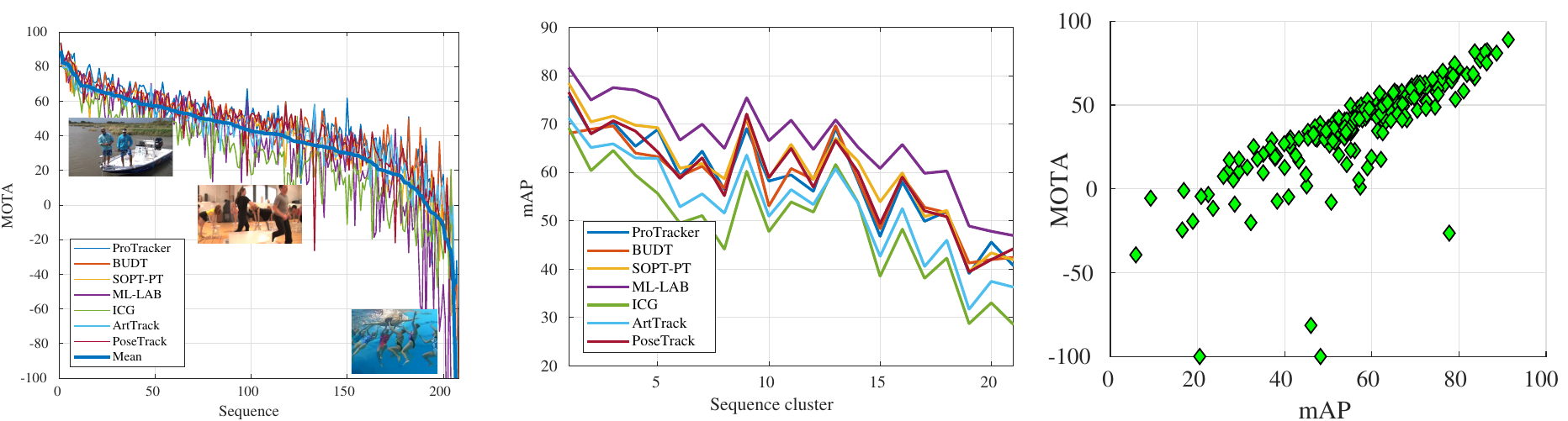}
  \vspace{-2mm}
  \caption{Sequences sorted by average MOTA (left). Pose estimation results sorted
    according to articulation complexity of the sequence (middle). Visualization
    of correlation between mAP and MOTA for each sequence (right). Note the outliers in
    right plot that correspond to sequences where pose estimation works well but
  tracking still fails. \vspace{-4mm}}
    \label{fig:mota:all_methods}
\end{figure*}

%% file: figure_easy_sequences.tex
\begin{figure*}[t]
	\centering
	%  \begin{tabular}{c c}
	%    \includegraphics[height=0.26\columnwidth]{images/annotations/annotation_sample1} & 
	%    \includegraphics[height=0.26\columnwidth]{images/annotations/annotation_sample2} 
	%    %\includegraphics[height=0.170\linewidth]{images/annotations/annotation_sample3} 
	%  \end{tabular}
	\includegraphics[height=0.134\linewidth]{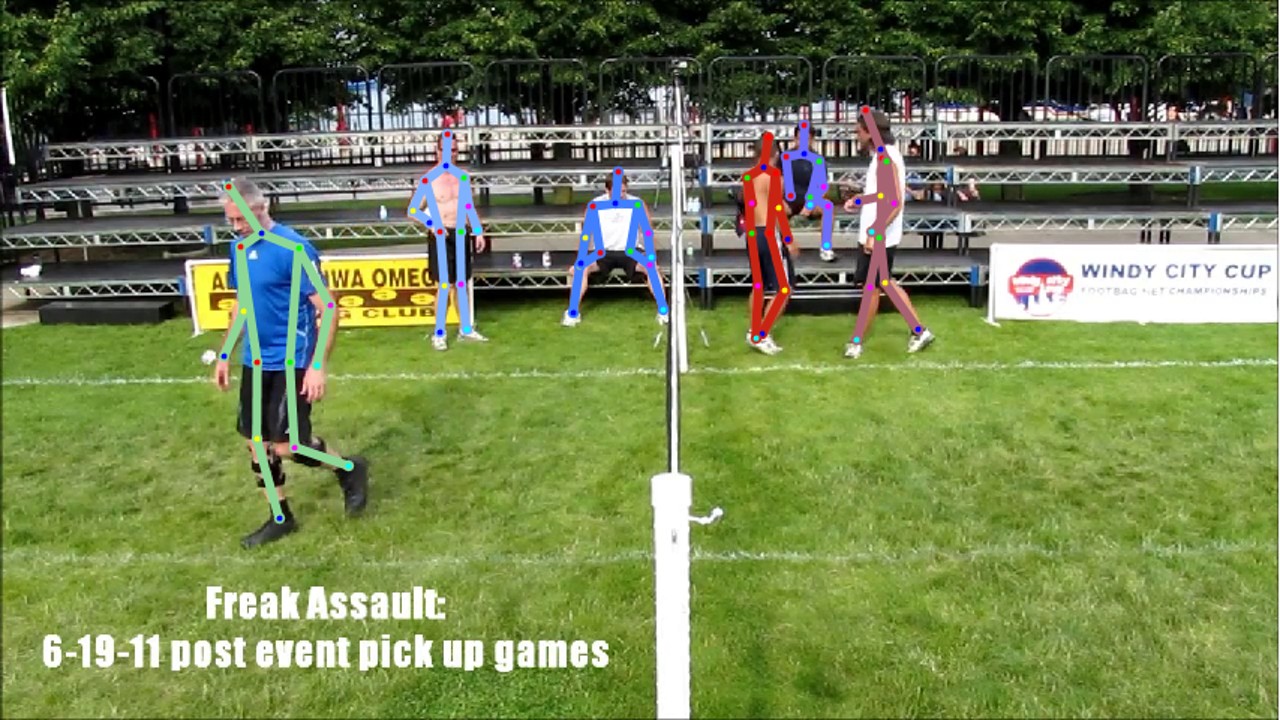} 
	\hfill
	\includegraphics[height=0.134\linewidth]{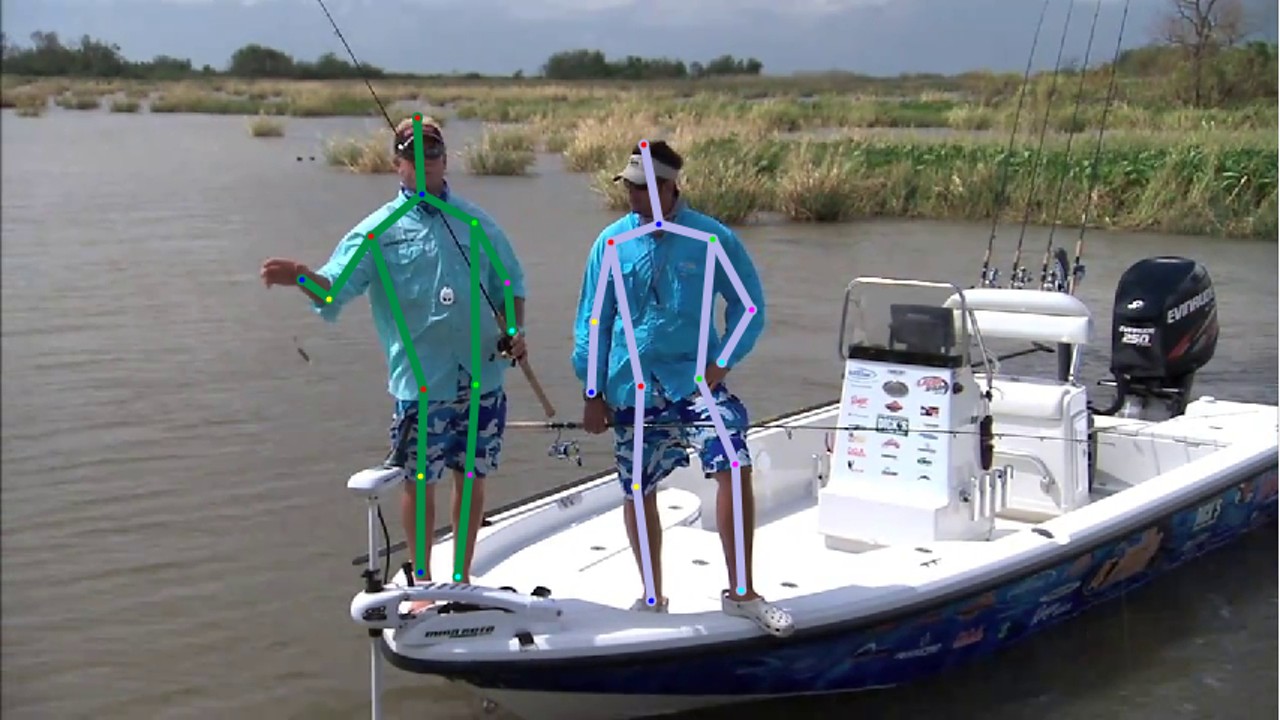} 
	\hfill
	\includegraphics[height=0.134\linewidth]{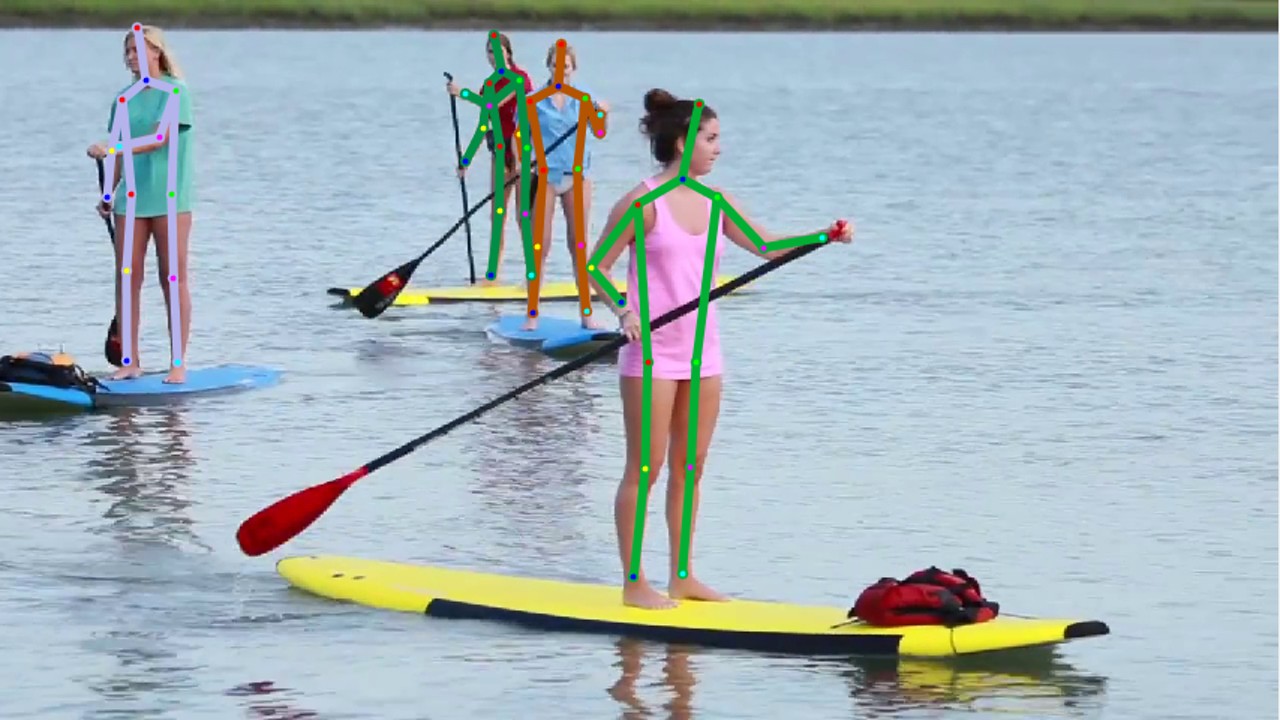} 
	\hfill
	\includegraphics[height=0.134\linewidth]{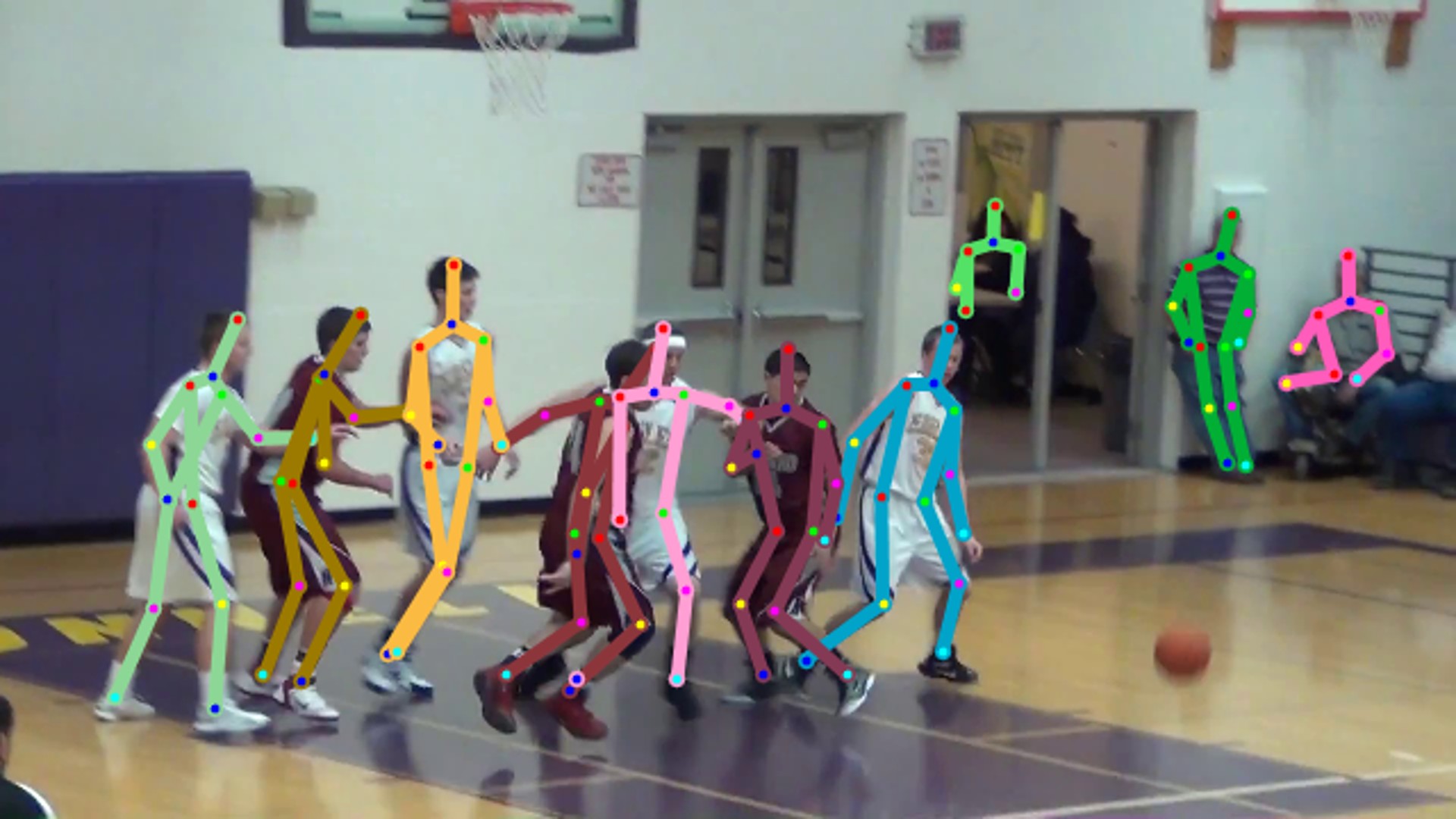} \\
	\vspace{-2mm}
	\caption{Selected frames from sample sequences with MOTA
          score above 75\% with predictions of our ArtTrack-baseline
          overlaid in each frame. See text for further description.}
    %\vspace{-2mm}
	\label{fig:easy_sequences}
\end{figure*}

% These videos typically feature
%           well separated individuals in upright standing
%           poses with minimal changes of body articulation over time
%           and no camera motion

%% file: figure_hard_sequences.tex
\newcommand{\cwidth}{0.29}
\begin{figure*}[t]
	\centering
	\begin{tabular}{c c c c}
		
		\includegraphics[height=\cwidth\columnwidth]{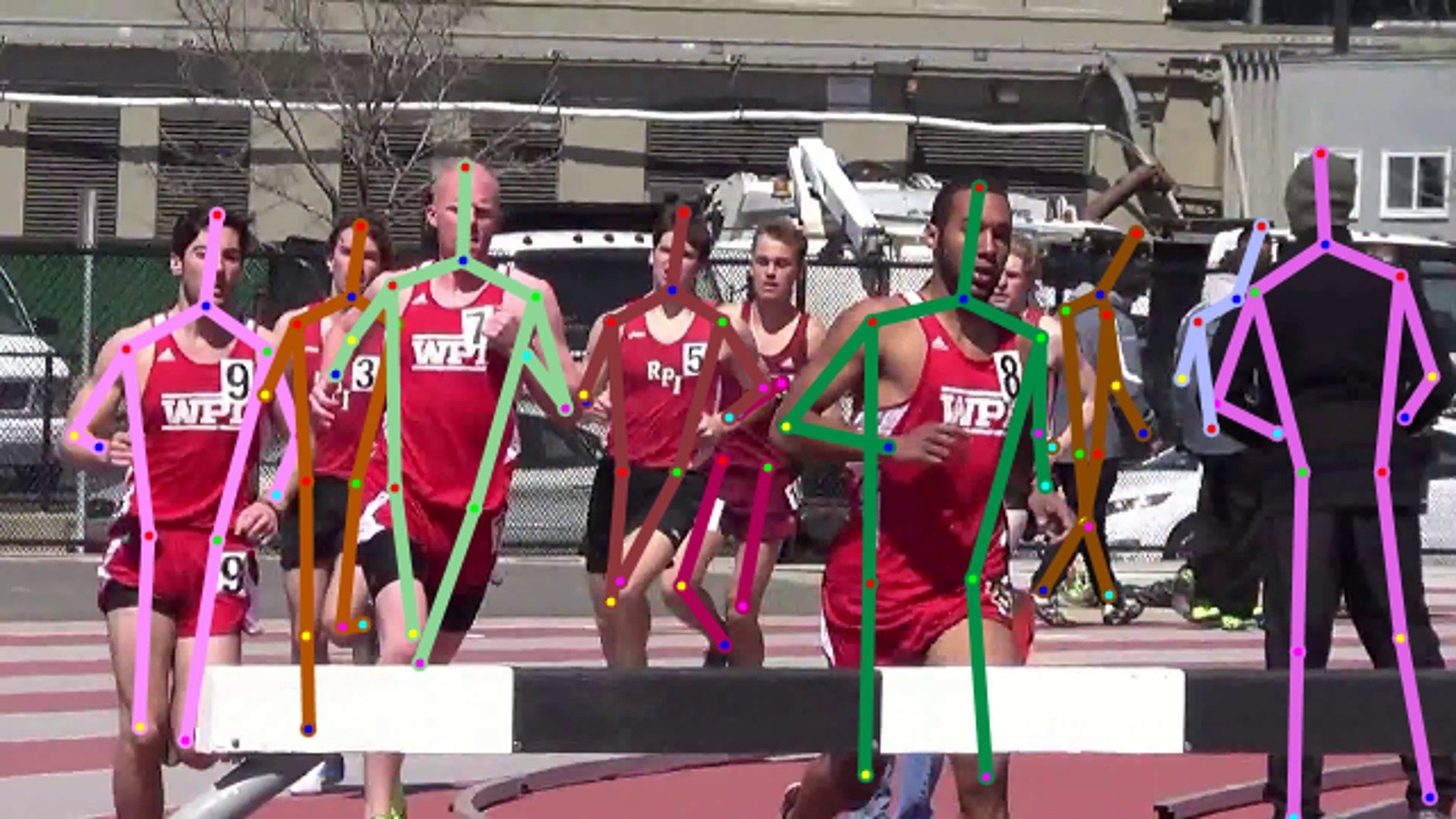} 
		&
		\includegraphics[height=\cwidth\columnwidth]{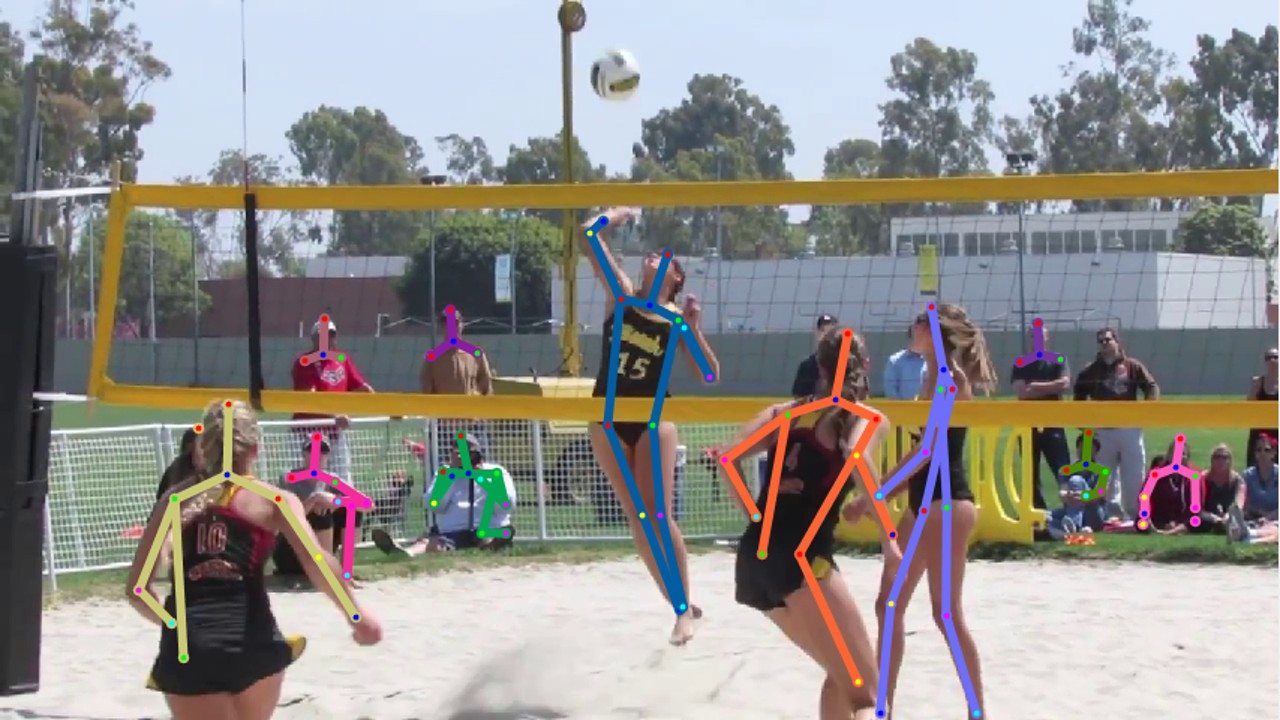} 
		&
		\includegraphics[height=\cwidth\columnwidth]{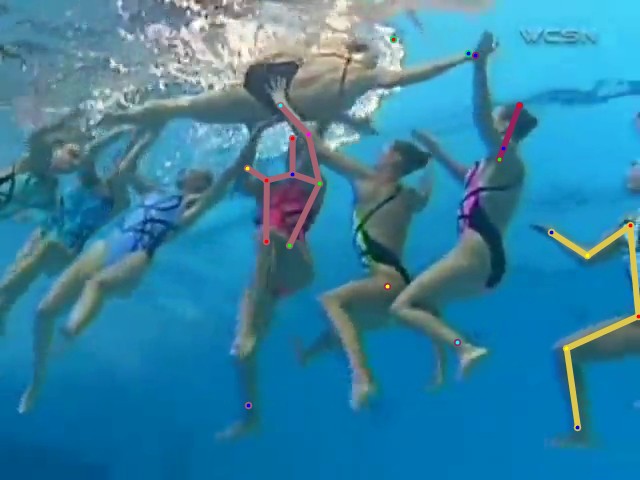} 	
		&
		\includegraphics[height=\cwidth\columnwidth]{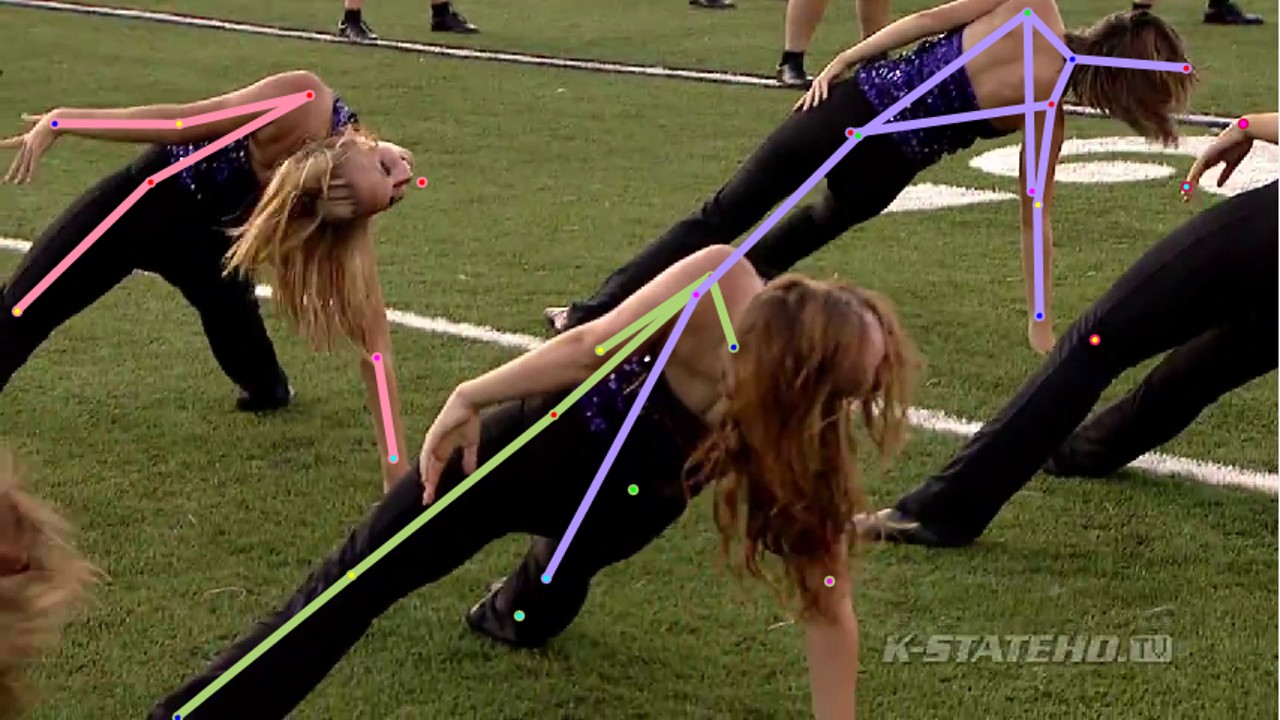} \\
		
		1 & 2 & 3 & 4 \\[.25em]
		
		\includegraphics[height=\cwidth\columnwidth]{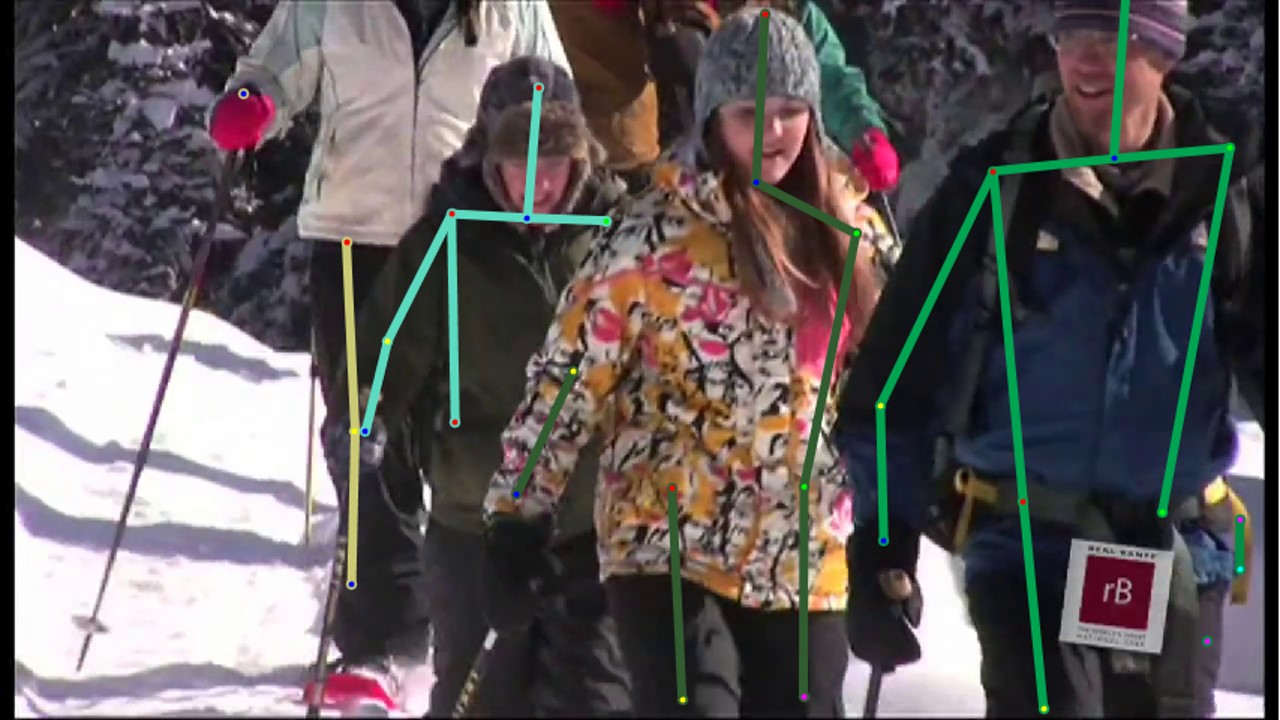} 
		&
		\includegraphics[height=\cwidth\columnwidth]{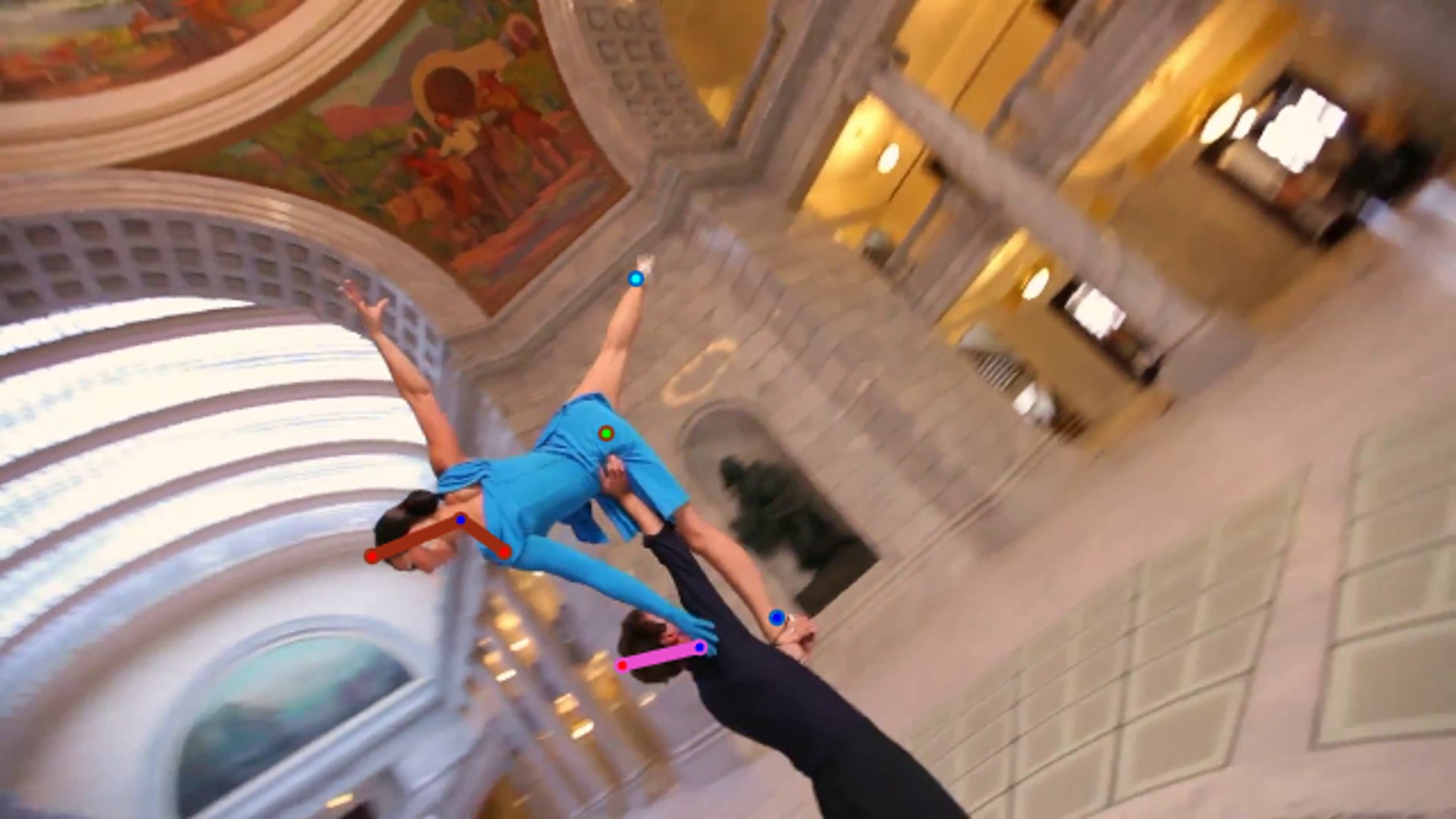} 
		&
		\includegraphics[height=\cwidth\columnwidth]{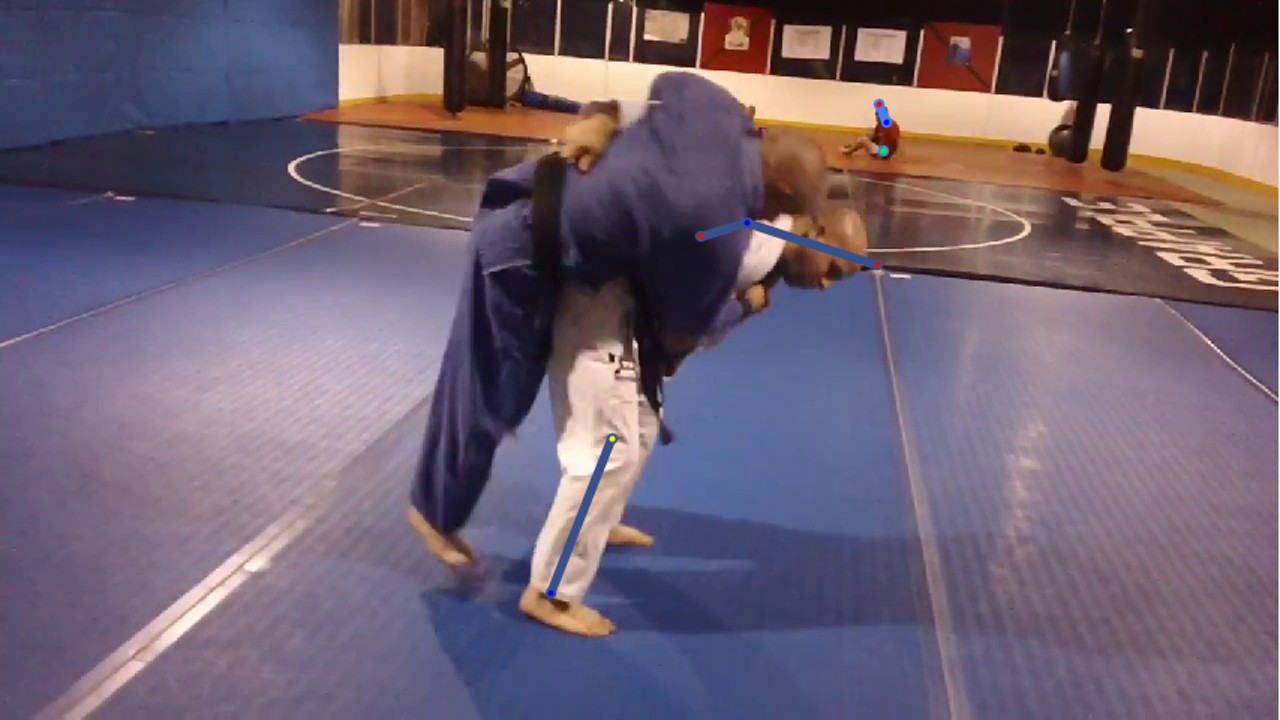} 
		&
		\includegraphics[height=\cwidth\columnwidth]{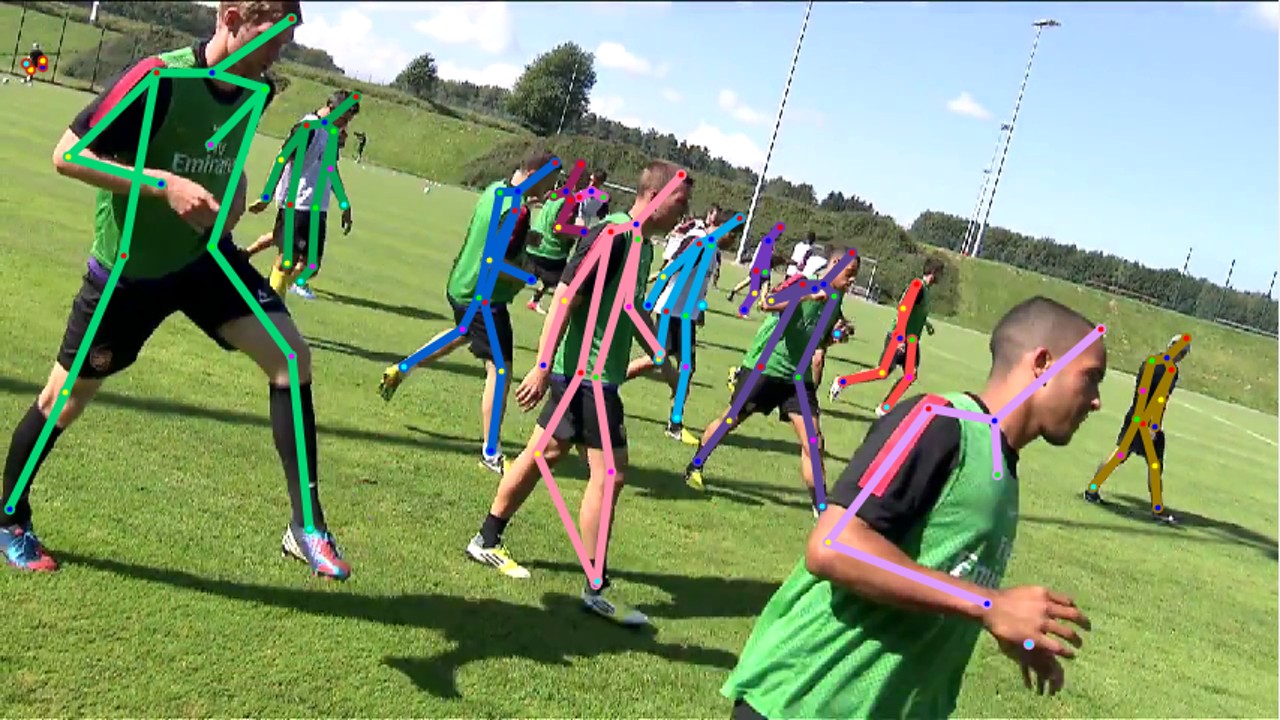} \\
		5 & 6 & 7 & 8 \\
		
	\end{tabular}
\vspace{-0.35cm}
	\caption{Selected frames from sample sequences with negative average MOTA
          score. The predictions of our ArtTrack-baseline are overlaid in each
          frame. Challenges for current methods in such sequences include crowds (images 3 and 8), extreme proximity of
          people to each other (7), rare poses (4 and 6) and strong camera motions (3, 5, 6, and
          8).}
     %\vspace{-2mm}     
	\label{fig:hard_sequences}
\end{figure*}

% \begin{figure*}[t]
% 	\centering
% 	\includegraphics[height=0.145\linewidth]{images/hard/000811_mpii_relpath_5sec_testsub_00000044.jpg} 
% 	\hfill
% 	\includegraphics[height=0.145\linewidth]{images/hard/09038_mpii_relpath_5sec_testsub_00000011.jpg} 
% 	\hfill
% 	\includegraphics[height=0.145\linewidth]{images/hard/09404_mpii_relpath_5sec_testsub_00000033.jpg} 
% 	\hfill
% 	\includegraphics[height=0.145\linewidth]{images/hard/15294_mpii_relpath_5sec_testsub_00000097.jpg} \\
% 	\hfill \\
	
% 	\includegraphics[height=0.136\linewidth]{images/hard/15375_mpii_relpath_5sec_testsub_00000069.jpg} 
% 	\hfill
% 	\includegraphics[height=0.136\linewidth]{images/hard/21086_mpii_relpath_5sec_testsub_00000046.jpg} 
% 	\hfill
% 	\includegraphics[height=0.136\linewidth]{images/hard/23965_mpii_relpath_5sec_testsub_00000048.jpg} 
% 	\hfill
% 	\includegraphics[height=0.136\linewidth]{images/16235_mpii_relpath_5sec_testsub_00000052.jpg} \\
	
% 	\caption{Frames from sequences with MOTA tracking accuracy below 0.}
% 	\label{fig:hard_sequences}
% \end{figure*}

%% file: figure_art_complexity.tex
% \begin{figure*}[!tbp]
%   \center
%   \begin{subfigure}[b]{0.30\textwidth}
%     \includegraphics[width=\textwidth]{figures/articulation_complexity_AP_plot}
%   \end{subfigure}
%   \hspace{2mm}
%   \begin{subfigure}[b]{0.30\textwidth}
%     \includegraphics[width=\textwidth]{figures/articulation_complexity_MOTA_plot}
%   \end{subfigure}
%   \hspace{2mm}
%   \begin{subfigure}[b]{0.32\textwidth}
%     \includegraphics[width=\textwidth]{figures/ap_mota}
%   \end{subfigure}
%   \vspace{-2mm}
%   \caption{Pose estimation (left) and pose tracking (middle) results sorted
%     according to articulation complexity of the sequence. The right plot visualizes
%     correlation between mAP and MOTA for each sequence. Note the outliers in
%     right plot that correspond to sequences where pose estimation works well but
%   tracking still fails. \vspace{-4mm}}
%     \label{fig:art_complexity}
% \end{figure*}

%% file: experiments.tex
\section{Dataset Analysis}
\label{sec:baselines}
%% We present the results of multi-person pose estimation accuracy on the
%% validation split in Table~\ref{tab:per-frame-ap-val}. We report the
%% results when training on our training set, MPII training set and on
%% the combined ours+MPII training set. Using our training set alone
%% achieves $55.5$\% mAP. Re-training on MPII training set improves
%% performance to $66.4$\% mAP. While our training set contains more
%% poses compared to MPII, multiple poses of the person are annotated in
%% each video, which results into a smaller number of the distinct
%% individuals compared to MPII dataset ($~2,400$ vs. $~4,000$).
%% %Therefore, the model trained on our
%% %video sequences struggles to capture the full variation in human
%% %appearances and poses.
%% It is important to note, however, that our dataset is the first to
%% contain such a large number of annotated video sequences, allowing for
%% training and analyzing the \emph{temporal} component of articulated
%% human pose tracking.
%% %\todo{Describe the trade-off between annotating more individual people vs. annotating the same person in video over time, argue that the strength of the dataset lies elsewhere}.
%% Re-training on the combined ours+MPII training set allows to improve
%% the performance further to $68.7$\% mAP, which shows both datasets'
%% complementarity.

%\myparagraph{Articulation complexity}
In order to better understand successes and failures of the current body pose
tracking approaches, we analyze their performance across the range of
sequences in the test set. To that end, for each sequence we compute an average
over MOTA scores obtained by each of the seven evaluated methods. Such average
score serves us as an estimate for the difficulty of the sequence for the
current computer vision approaches. We then rank the sequences by the average
MOTA. The resulting ranking is shown in Fig.~\ref{fig:mota:all_methods} (left) along
with the original MOTA scores of each of the approaches.
%
% We use MOTA metric to characterize the
% performance of each approach.
% We proceed by sorting the test sequences according
% to the averaged per-sequence MOTA.
%Sorting is performed in descending order from easiest to hardest
%sequences. Results are shown in .
%
First, we observe that all methods
perform similarly well on easy
sequences. Fig.~\ref{fig:easy_sequences} shows a few easy sequences
with an average MOTA above $75\%$. Visual analysis
reveals that easy sequences typically contain significantly separated
individuals in upright standing poses with minimal changes of body
articulation over time and no camera motion. Tracking accuracy drops
 with the increased complexity of video
sequences. Fig.~\ref{fig:hard_sequences} shows a few hard sequences
with average MOTA accuracy below $0$. 
These sequences typically include strongly overlapping people, and fast motions of
people and camera.

%% Fig.~\ref{fig:art_complexity} demonstrates how the performance of all
%% submission varies \wrt the complexity of body pose articulation. The
%% articulation of each video sequence is calculated as the difference of
%% body poses in that video with the mean pose of the dataset.

%% %\subsection{Analysis}
%% Figure~\ref{fig:mota:all_methods} shows MOTA scores of the sequences
%% sorted by the MOTA averaged across all submitted methods, from the
%% most difficult to the easiest of sequences. Superimposed are MOTA
%% scores of each method. An immediate observation is that various
%% methods generally agree on the difficulty of different sequences.

%\begin{figure}[t]
%		\includegraphics[trim=0 0 0 0,clip,width=1.0\linewidth]{figures/ap_mota} \\
%		\caption{ Correlation between AP and MOTA metrics per sequence. }
%	\label{fig:mota_ap_correlation}
%\end{figure}

We further analyze how tracking and pose estimation accuracy are affected by
pose complexity. As a measure for the pose
complexity of a sequence we employ an average deviation of each pose in a sequence
from the mean pose. 
% First, we compute per-joint Euclidean distance between
% each body pose and the mean body pose, and then average all distances per
% sequence. 
The computed complexity score is used to sort video sequences from low to high pose
complexity and average mAP is reported for each
sequence. The result of this evaluation is shown in
Fig.~\ref{fig:mota:all_methods} (middle). For visualization purposes, we partition the sorted video
sequences into bins of size 10 based on pose complexity score and report average
mAP for each bin. We observe that both body pose estimation and tracking
performance significantly decrease with the increased pose complexity.
% This is
% to be expected as more complex poses are typically not well represented in the
% training set and thus are harder to reliably detect and track at the test
% time.
Fig.~\ref{fig:mota:all_methods} (right) shows a plot that highlights correlation
between mAP and MOTA of the same sequence. We use the mean performance of all methods in this visualization. Note that in most cases more accurate
pose estimation reflected by higher mAP indeed corresponds to higher
MOTA. However, it is instructive to look at sequences where poses are estimated
accurately (mAP is high), yet tracking results are particularly poor (MOTA near
zero). One of such sequences is shown in Fig.~\ref{fig:hard_sequences} (8). This
sequence features a large number of people and fast camera movement that is likely
confusing simple frame-to-frame association tracking of the evaluated
approaches. Please see supplemental material for additional examples and
analyses of challenging sequences.

%% file: conclusion.tex
\section{Conclusion}
\label{sec:conclusion}
In this paper we proposed a new benchmark for human pose estimation and articulated tracking that is significantly larger and more diverse in terms of data variability and complexity compared to existing pose tracking benchmarks. Our benchmark enables objective comparison of different approaches for articulated people tracking in realistic scenes. We have set up an online evaluation server that permits evaluation on a held-out test set, and have measures in place to limit overfitting on the dataset. Finally, we conducted a rigorous survey of the state of the art. Due to the scale and complexity of the benchmark, most existing methods build on combinations of proven components: people detection, single-person pose estimation, and tracking based on simple association between neighboring frames. Our analysis shows that current methods perform well on easy sequences with well separated upright people, but are severely challenged in the presence of fast camera motions and complex articulations. Addressing these challenges remains an important direction for the future work.

% is a promising direction for the future work. Another direction for future work is to improve efficiency of human labeling by leveraging weakly-supervised and active learning methods.

% focus on weakly-supervised and active learning methods to make more efficient use of the human annotators, and on other means of assembling large
% None of the top methods leverage end-to-end methods neither on the frame level to jointly infer poses of multiple people, nor on the video level to predict entire trajectories of people.
% % This is likely to happen in the future and will be better positioned to address harder parts of the dataset.
% % We identified the challenging subsets of the dataset by ranking sequences according to the performance of the current methods.
% End-to-end training might be limited at the moment by the size of the available training set. Future work should focus on weakly-supervised and active learning methods to make more efficient use of the human annotators, and on other means of

%% file: acknowledgements.tex
\myparagraph{Acknowledgements.}
UI and JG have been supported by the DFG project GA 1927/5-1 (FOR 2535) and the ERC Starting Grant ARCA (677650).

%% file: posetrack_paper.bbl
\begin{thebibliography}{10}\itemsep=-1pt

\bibitem{andriluka14cvpr}
M.~Andriluka, L.~Pishchulin, P.~Gehler, and B.~Schiele.
\newblock 2{D} human pose estimation: New benchmark and state of the art
  analysis.
\newblock In {\em CVPR}, 2014.

\bibitem{bulat2016human}
A.~Bulat and G.~Tzimiropoulos.
\newblock Human pose estimation via convolutional part heatmap regression.
\newblock In {\em ECCV}, 2016.

\bibitem{cao16arxiv}
Z.~Cao, T.~Simon, S.-E. Wei, and Y.~Sheikh.
\newblock Realtime multi-person 2{D} pose estimation using part affinity
  fields.
\newblock In {\em CVPR}, 2017.

\bibitem{carreira16cvpr}
J.~Carreira, P.~Agrawal, K.~Fragkiadaki, and J.~Malik.
\newblock Human pose estimation with iterative error feedback.
\newblock In {\em CVPR}, 2016.

\bibitem{Carreira_2017_CVPR}
J.~Carreira and A.~Zisserman.
\newblock Quo vadis, action recognition? a new model and the kinetics dataset.
\newblock In {\em CVPR}, 2017.

\bibitem{Charles16}
J.~Charles, T.~Pfister, D.~Magee, and A.~Hogg, D.~Zisserman.
\newblock Personalizing human video pose estimation.
\newblock In {\em CVPR}, 2016.

\bibitem{chen2016deeplab}
L.-C. Chen, G.~Papandreou, I.~Kokkinos, K.~Murphy, and A.~L. Yuille.
\newblock Deeplab: Semantic image segmentation with deep convolutional nets,
  atrous convolution, and fully connected crfs.
\newblock {\em arXiv preprint arXiv:1606.00915}, 2016.

\bibitem{Choi:2015:ICCV}
W.~Choi.
\newblock Near-online multi-target tracking with aggregated local flow
  descriptor.
\newblock In {\em ICCV 2015}.

\bibitem{dantone13cvpr}
M.~Dantone, J.~Gall, C.~Leistner, and L.~V. Gool.
\newblock Human pose estimation using body parts dependent joint regressors.
\newblock In {\em CVPR}, 2013.

\bibitem{eichner10eccv}
M.~Eichner and V.~Ferrari.
\newblock We are family: Joint pose estimation of multiple persons.
\newblock In {\em ECCV}, 2010.

\bibitem{ProTracker}
R.~Girdhar, G.~Gkioxari, L.~Torresani, D.~Ramanan, M.~Paluri, and D.~Tran.
\newblock Simple, efficient and effective keypoint tracking.
\newblock In {\em ICCV PoseTrack Workshop}, 2017.

\bibitem{gkioxari16eccv}
G.~Gkioxari, A.~Toshev, and N.~Jaitly.
\newblock Chained predictions using convolutional neural networks.
\newblock In {\em ECCV}, 2016.

\bibitem{he17iccv}
K.~He, G.~Gkioxari, P.~Dollár, and R.~Girshick.
\newblock Mask r-cnn.
\newblock In {\em ICCV}, 2017.

\bibitem{he2015deep}
K.~He, X.~Zhang, S.~Ren, and J.~Sun.
\newblock Deep residual learning for image recognition.
\newblock {\em arXiv preprint arXiv:1512.03385}, 2015.

\bibitem{hu2016bottom}
P.~Hu and D.~Ramanan.
\newblock Bottom-up and top-down reasoning with hierarchical rectified
  gaussians.
\newblock In {\em CVPR}, {2016}.

\bibitem{huang2016speed}
J.~Huang, V.~Rathod, C.~Sun, M.~Zhu, A.~Korattikara, A.~Fathi, I.~Fischer,
  Z.~Wojna, Y.~Song, S.~Guadarrama, et~al.
\newblock Speed/accuracy trade-offs for modern convolutional object detectors.
\newblock {\em arXiv preprint arXiv:1611.10012}, 2016.

\bibitem{insafutdinov17cvpr}
E.~Insafutdinov, M.~Andriluka, L.~Pishchulin, S.~Tang, E.~Levinkov, B.~Andres,
  and B.~Schiele.
\newblock Arttrack: Articulated multi-person tracking in the wild.
\newblock In {\em CVPR}, 2017.

\bibitem{insafutdinov16eccv}
E.~Insafutdinov, L.~Pishchulin, B.~Andres, M.~Andriluka, and B.~Schiele.
\newblock Deepercut: A deeper, stronger, and faster multi-person pose
  estimation model.
\newblock In {\em ECCV}, 2016.

\bibitem{Ionescu:2014:H36}
C.~Ionescu, D.~Papava, V.~Olaru, and C.~Sminchisescu.
\newblock {Human3.6M: Large Scale Datasets and Predictive Methods for 3D Human
  Sensing in Natural Environments}.
\newblock {\em PAMI}, 2014.

\bibitem{Iqbal_ECCVw2016}
U.~Iqbal and J.~Gall.
\newblock Multi-person pose estimation with local joint-to-person associations.
\newblock In {\em ECCVw}, 2016.

\bibitem{iqbal17fg}
U.~Iqbal, M.~Garbade, and J.~Gall.
\newblock Pose for action - action for pose.
\newblock In {\em FG}, 2017.

\bibitem{Iqbal:2017:CVPR}
U.~Iqbal, A.~Milan, and J.~Gall.
\newblock Pose{T}rack: {J}oint multi-person pose estimation and tracking.
\newblock In {\em CVPR}, 2017.

\bibitem{Jhuang:ICCV:2013}
H.~Jhuang, J.~Gall, S.~Zuffi, C.~Schmid, and M.~J. Black.
\newblock Towards understanding action recognition.
\newblock In {\em ICCV}, 2013.

\bibitem{BUDT}
S.~Jin, X.~Ma, Z.~Han, Y.~Wu, W.~Yang, W.~Liu, C.~Qian, and W.~Ouyang.
\newblock Towards multi-person pose tracking: Bottom-up and top-down methods.
\newblock In {\em ICCV PoseTrack Workshop}, 2017.

\bibitem{johnson10bmvc}
S.~Johnson and M.~Everingham.
\newblock Clustered pose and nonlinear appearance models for human pose
  estimation.
\newblock In {\em BMVC}, 2010.

\bibitem{johnson11cvpr}
S.~Johnson and M.~Everingham.
\newblock {Learning Effective Human Pose Estimation from Inaccurate
  Annotation}.
\newblock In {\em CVPR}, 2011.

\bibitem{Levinkov_2017_CVPR}
E.~Levinkov, J.~Uhrig, S.~Tang, M.~Omran, E.~Insafutdinov, A.~Kirillov,
  C.~Rother, T.~Brox, B.~Schiele, and B.~Andres.
\newblock Joint graph decomposition and node labeling: Problem, algorithms,
  applications.
\newblock In {\em CVPR}, 2017.

\bibitem{lin14eccv}
T.-Y. Lin, M.~Maire, S.~Belongie, J.~Hays, P.~Perona, D.~Ramanan, P.~Dollár,
  and C.~L. Zitnick.
\newblock Microsoft coco: Common objects in context.
\newblock In {\em ECCV}, 2014.

\bibitem{liu2016ssd}
W.~Liu, D.~Anguelov, D.~Erhan, C.~Szegedy, S.~Reed, C.-Y. Fu, and A.~C. Berg.
\newblock Ssd: Single shot multibox detector.
\newblock In {\em European conference on computer vision}, pages 21--37.
  Springer, 2016.

\bibitem{Milan:2016:MOT16}
A.~Milan, L.~Leal-Taix\'{e}, I.~Reid, S.~Roth, and K.~Schindler.
\newblock Mot16: A benchmark for multi-object tracking.
\newblock {\em arXiv:1603.00831 [cs]}, 2016.

\bibitem{newell16eccv}
A.~Newell, K.~Yang, and J.~Deng.
\newblock Stacked hourglass networks for human pose estimation.
\newblock In {\em ECCV}, 2016.

\bibitem{papandreou17arxiv}
G.~Papandreou, T.~Zhu, N.~Kanazawa, A.~Toshev, J.~Tompson, C.~Bregler, and
  K.~Murphy.
\newblock Towards accurate multi-person pose estimation in the wild.
\newblock In {\em CVPR}, 2017.

\bibitem{ICG}
C.~Payer, T.~Neff, H.~Bischof, M.~Urschler, and D.~Stern.
\newblock Simultaneous multi-person detection and single-person pose estimation
  with a single heatmap regression network.
\newblock In {\em ICCV PoseTrack Workshop}, 2017.

\bibitem{Pfister15}
T.~Pfister, J.~Charles, and A.~Zisserman.
\newblock Flowing convnets for human pose estimation in videos.
\newblock In {\em ICCV}, 2015.

\bibitem{pishchulin16cvpr}
L.~Pishchulin, E.~Insafutdinov, S.~Tang, B.~Andres, M.~Andriluka, P.~Gehler,
  and B.~Schiele.
\newblock Deepcut: Joint subset partition and labeling for multi person pose
  estimation.
\newblock In {\em CVPR}, 2016.

\bibitem{rafi2016bmvc}
U.~Rafi, I.Kostrikov, J.~Gall, and B.~Leibe.
\newblock An efficient convolutional network for human pose estimation.
\newblock In {\em BMVC}, 2016.

\bibitem{ren2015faster}
S.~Ren, K.~He, R.~Girshick, and J.~Sun.
\newblock Faster {R-CNN}: {T}owards real-time object detection with region
  proposal networks.
\newblock In {\em NIPS}, pages 91--99, 2015.

\bibitem{sapp13cvpr}
B.~Sapp and B.~Taskar.
\newblock Multimodal decomposable models for human pose estimation.
\newblock In {\em CVPR}, 2013.

\bibitem{sapp11cvpr}
B.~Sapp, D.~Weiss, and B.~Taskar.
\newblock Parsing human motion with stretchable models.
\newblock In {\em CVPR}, 2011.

\bibitem{Sigal:2010:HSV}
L.~Sigal, A.~Balan, and M.~J. Black.
\newblock Humaneva: Synchronized video and motion capture dataset and baseline
  algorithm for evaluation of articulated human motion.
\newblock {\em International Journal of Computer Vision}, 87, 2010.

\bibitem{Simonyan14c}
K.~Simonyan and A.~Zisserman.
\newblock Very deep convolutional networks for large-scale image recognition.
\newblock {\em CoRR}, 2014.

\bibitem{szegedy2017inception}
C.~Szegedy, S.~Ioffe, V.~Vanhoucke, and A.~A. Alemi.
\newblock Inception-v4, inception-resnet and the impact of residual connections
  on learning.
\newblock In {\em AAAI}, pages 4278--4284, 2017.

\bibitem{SOPT-PT}
TODO.
\newblock Towards realtime 2d pose tracking: A simple online pose tracker.
\newblock In {\em ICCV PoseTrack Workshop}, 2017.

\bibitem{Tompson:2015:EOL}
J.~Tompson, R.~Goroshin, A.~Jain, Y.~LeCun, and C.~Bregler.
\newblock Efficient object localization using convolutional networks.
\newblock In {\em CVPR}, 2015.

\bibitem{tompson14nips}
J.~Tompson, A.~Jain, Y.~LeCun, and C.~Bregler.
\newblock Joint training of a convolutional network and a graphical model for
  human pose estimation.
\newblock In {\em NIPS}, 2014.

\bibitem{Toshev:2014:DHP}
A.~Toshev and C.~Szegedy.
\newblock Deeppose: Human pose estimation via deep neural networks.
\newblock In {\em CVPR}, 2014.

\bibitem{varol17b}
G.~Varol, J.~Romero, X.~Martin, N.~Mahmood, M.~J. Black, I.~Laptev, and
  C.~Schmid.
\newblock {Learning from Synthetic Humans}.
\newblock In {\em CVPR}, 2017.

\bibitem{vodrick12ijcv}
C.~Vondrick, D.~Patterson, and D.~Ramanan.
\newblock Efficiently scaling up crowdsourced video annotation.
\newblock {\em IJCV'12}.

\bibitem{wei16cvpr}
S.-E. Wei, V.~Ramakrishna, T.~Kanade, and Y.~Sheikh.
\newblock Convolutional pose machines.
\newblock In {\em CVPR}, 2016.

\bibitem{Yang:2012:CVPR}
B.~Yang and R.~Nevatia.
\newblock An online learned {CRF} model for multi-target tracking.
\newblock In {\em CVPR 2012}, pages 2034--2041.

\bibitem{zhang2013actemes}
W.~Zhang, M.~Zhu, and K.~G. Derpanis.
\newblock From actemes to action: A strongly-supervised representation for
  detailed action understanding.
\newblock In {\em CVPR}, 2013.

\bibitem{ML-LAB}
X.~Zhu, Y.~Jiang, and Z.~Luo.
\newblock Multi-person pose estimation for posetrack with enhanced part
  affinity fields.
\newblock In {\em ICCV PoseTrack Workshop}, 2017.

\end{thebibliography}
